%% file: main.tex
\documentclass{article}

    \usepackage[preprint]{neurips_data_2024}

\usepackage[utf8]{inputenc} 
\usepackage[T1]{fontenc}    
\usepackage{hyperref}       
\usepackage{url}            
\usepackage{booktabs}       
\usepackage{amsfonts}       
\usepackage{nicefrac}       
\usepackage{microtype}      
\usepackage{xcolor}         
\usepackage{graphicx}
\usepackage{multirow}
\usepackage{algorithm}
\usepackage{algpseudocode}
\usepackage{makecell}
\usepackage{fixltx2e}
\usepackage{tabularx}

\usepackage{enumitem}

\title{DStruct2Design: Data and Benchmarks for Data Structure Driven Generative Floor Plan Design}

\author{%
  Zhi Hao Luo \\
  MILA, Polytechnique Montréal\\
  \texttt{luozhiha@mila.quebec} \\
  \And
  Luis Lara \\
  MILA \\
  \texttt{luis.lara@mila.quebec} \\
  \And
  Ge Ya Luo \\
  MILA, Université de Montréal \\
  \texttt{xugeya@mila.quebec} \\
  \And
  Florian Golemo \\
  MILA \\
  \texttt{fgolemo@gmail.com} \\
  \And
  Christopher Beckham \\
  MILA \\
  \texttt{first.last@mila.quebec} \\
  \And
  Christopher Pal \\
  MILA, Canada CIFAR AI Chair\\
  \texttt{christopher.pal@polymtl.ca} \\
}

\begin{document}

\maketitle

\begin{abstract}
  Text conditioned generative models for images have yielded impressive results. Text conditioned floorplan generation as a special type of raster image generation task also  received particular attention. 
  However there are many use cases in floorplan generation where numerical properties of the generated result are more important than the aesthetics. For instance, one might want to specify sizes for certain rooms in a floorplan and compare the generated floorplan with given specifications. Current approaches, datasets and commonly used evaluations do not support these kinds of constraints. As such, an attractive strategy is to generate an intermediate data structure that contains numerical properties of a floorplan which can be used to generate the final floorplan image. To explore this setting we (1) construct a new dataset for this data-structure to data-structure formulation of floorplan generation using two popular image based floorplan datasets RPLAN and ProcTHOR-10k, and provide the tools to convert further procedurally generated ProcTHOR floorplan data into our format. 
  (2) We explore the task of floorplan generation given a partial or complete set of constraints and we design a series of metrics and benchmarks to enable evaluating how well samples generated from models respect the constraints. 
  (3) We create multiple baselines by finetuning a large language model (LLM), Llama3, and demonstrate the feasibility of using floorplan data structure conditioned LLMs for the problem of floorplan generation respecting numerical constraints.
  We hope that our new datasets and benchmarks will encourage further research on different ways to improve the performance of LLMs and other generative modelling techniques for generating designs where quantitative constraints are only partially specified, but must be respected.\footnote{code for our project is available on github: https://github.com/plstory/DS2D}
\end{abstract}

\section{Introduction}
\label{sec:intro}
\input{sections/introduction}

\section{Related Work}
\label{sec:related}
\input{sections/related_work}

\section{Our Text-Based DStruct2Design Floorplan Dataset}
\label{sec:method}
\input{sections/dataset}

\section{Our DStruct2Design (DS2D) LLM}
\label{sec:llm}
\input{sections/llm}

\section{Our Metrics and Benchmarks}
\label{sec:metrics}
\input{sections/metrics}

\section{Experiments}
\label{sec:experiments}
\input{sections/experiments}

\section{Results}
\label{sec:results}
\input{sections/Results}

\vspace{-0.7cm}
\section{Conclusions}\label{sec:conclusion}
\input{sections/conclusion}

We have motivated the need for new datasets and benchmarks for real-world use case scenarios for floorplan generation. We have developed and explored a Llama 3 based LLM for the problem and it yields SOTA results on the previous compatibility based evaluation and along with our proposed metrics, it highlights different use cases scenarios where improvements are possible. 
In particular we see from Table 7 that when conditioning on bubble diagrams and room area information this model struggles to respect the bubble diagram constraints. From Table 5 we see that the model does well, but is far from perfect at generating rooms with correct areas as computed from the generated polygons. We also see that the biggest weakness of this otherwise SOTA model is linked to issues with generating overlapping rooms.   

We hope that our data, simulated data generation procedure and benchmarks will stimulate further developments for this new problem formulation for floorplan generation. 

\bibliographystyle{unsrtnat}

\bibliography{main}


\input{sections/appendix}
\end{document}

%% file: sections/introduction.tex
Generative modelling has the potential to accelerate and improve design tasks but in order for them to achieve widespread use by real-world practitioners they need to be flexible and exhibit a sufficiently high degree of controllability. For instance, in floorplan generation users will want to be able to specify a set of constraints detailing how rooms should be connected and what their dimensions are. The resulting model should also be flexible in the sense that it performs equally well on as few or as many constraints as needed. Just as there are many possible constraints that the user can specify, there also needs to be a rigorously-defined set of evaluation metrics to probe how well the model performs (or underperforms) on them, which we contribute in this work.

Most existing approaches to floorplan-based generative models focus on 2D top-down renders. These methods typically specify room connectivity via bubble diagrams as well as other information such as room geometries. However, these methods fall short in various areas, for instance some produce raster-based renderings of the floorplan (e.g. vision-based generative models) which make it difficult to directly access metadata. Others may only support the specification of adjacencies instead of also supporting room geometry, limiting the controllability of the model. In this work, we instead opt to represent generated floorplans as a JSON-based data structure specifying all the required metadata of the floorplan as well as specifying each room as a list of polygons (similar to \cite{shabani2022housediffusion}). These floorplans are a result of consolidating two existing large floorplan datasets, namely RPLAN and ProcTHOR. Furthermore, we validate the use of such a dataset by training an LLM to produce such a structure conditioned on constraints specified in natural language. Therefore, our contributions are as follows:


\begin{enumerate}[leftmargin=10pt]
\item We contribute a unified floorplan dataset by merging existing popular datasets (RPLAN and ProcTHOR) and postprocessing them into a format which is amenable to language-based generative models which take as input and output structured text.
\item We design an array of metrics to benchmark the performance of the generated floorplans, in particular how well they respect the constraints set by the user.
\item We train and evaluate an LLM on our task setting to demonstrate the feasibility of our proposed problem formulation.
\end{enumerate}



%% file: sections/related_work.tex

\textbf{Datasets.} One of the most commonly used datasets for floorplan generation is RPLAN \citep{rplan}. Since we make use of the dataset in this work, we defer its description to Section \ref{sec:data}. LIFULL \citep{lifull} is an extremely large-scale dataset consisting of floorplans sourced from a Japanese realtor company. Floorplans are provided in raster format, however a subset of vectorized versions exist \citep{liu2017raster}. One common shortcoming is that there is a lack of well-annotated floorplan data (e.g. vectorised) which is amenable to training. Architext \citep{galanos2023architext} address this issue by generating synthetic but diverse floorplans via a CAD script for Rhinoceros 3D. In a similar vein, we leverage ProcTHOR \citep{procthor} (also used and described in more detail in Section \ref{sec:data}) which is a fully open source simulator of buildings (and by extension floorplans). Unlike Rhinoceros 3D this simulator is completely open source.


\textbf{Generative Models.} \textit{House-GAN} and \textit{House-GAN++} \citep{nauata2020house,nauata2021house} are a family of GAN-based methods which learn to generate floorplans using convolutional graph-based networks. Notably however this line of work only conditions on a bubble graph. \textit{FloorplanGAN} \citep{FPGAN} proposes a self-attention-based GAN which takes as input (for each room) room centers, desired areas (as a relative proportions), as well as room type for each room. Because the GAN is tasked with also refining the initial input constraints, the output may not respect them. While the use of a differentiable rasteriser to compute losses in pixel space opens up possibilities, the core method does not appear to support a partial specification of constraints or polygons. \textit{HouseDiffusion} \citep{shabani2023housediffusion} is a diffusion-based method which directly predicts a list of polygons for each room, utilising a transformer architecture which also conditions on a bubble diagram.\footnote{While this method could in principle be conditioned on an initial polygonal room specification (via partial noising and subsequent denoising), from the point of view of pure inference (that is, starting from pure noise) \textit{HouseDiffusion}'s only form of conditioning is via a bubble diagram.}Lastly, \textit{ArchiText} \citep{galanos2023architext} also leverages LLMs to generate output floorplans but it appears the prompts are only limited to natural language descriptions rather than geometry.



\textbf{3D Scene Generation.} Recently, full 3D scene generation methods have  shown impressive results.
AnyHome \citep{wen2023anyhome} and Holodeck \citep{yang2024holodeck} are able to generate floor plans, windows \& doors, furniture and meaningful placement in 3D all from a query like "a 1b1b apartment". 
In our method, we focus only on the floor plan aspect, but we allow for specification of room dimensions and areas as well as total floor plan area, which neither method does. Since our proposed dataset is partially comprised of ProcTHOR (a 3D-based floorplan simulator), in principle a future version of our proposed dataset could support the placement of furniture and other props.
An overview of how our method compares to the others can be found in Tab.\ref{tab:method-comparison}.

{\renewcommand{\arraystretch}{1.2}%
\begin{table}[hbt!]
\small\begin{tabular}{@{}lllcc@{}}
\toprule
\textbf{Method}                                                                  & \textbf{Input}                                                            & \textbf{Output}                                                              & \multicolumn{1}{l}{\begin{tabular}[c]{@{}l@{}}\textbf{Can specify} \\ \textbf{Geometry?}\end{tabular}} & \multicolumn{1}{l}{\begin{tabular}[c]{@{}l@{}}\textbf{Absolute} \\ \textbf{Coordinates?}\end{tabular}} \\ \midrule
\begin{tabular}[c]{@{}l@{}}House-GAN,\\ House-GAN++,\\ HouseDiffusion\end{tabular} & \begin{tabular}[c]{@{}l@{}}Bubble diagram\\ (room adjacency)\end{tabular} & Room masks                                                                   & no                                                                                   & no                                                                                   \\
ArchiText                                                                        & Natural Language                                                          & \begin{tabular}[c]{@{}l@{}}Room polygon coords\end{tabular}          & no                                                                                   & yes                                                                                  \\
FloorplanGAN                                                                     & Fixed-Length Tensor                                                       & Room masks                                                                   & yes                                                                                  & no                                                                                   \\
\begin{tabular}[c]{@{}l@{}}AnyHome,\\ Holodeck\end{tabular}                       & Natural Language                                                          & \begin{tabular}[c]{@{}l@{}}Furnished 3D models\end{tabular}         & no                                                                                   & yes                                                                                  \\
\textbf{Ours}                                                                    & \textbf{Structured Language}                                              & \textbf{\begin{tabular}[c]{@{}l@{}}Room polygon coords\end{tabular}} & \textbf{yes}                                                                         & \textbf{yes}                                                                         \\ \bottomrule
\end{tabular}
\caption{\textbf{Comparison to related works.} Our method focuses on generation of floor plans as opposed to full 3D scenes, and it allows the user to specify individual room geometry (height, width, and/or area) and global apartment geometry in meters and sq. meters.}
\label{tab:method-comparison}
\end{table}}
\vspace{-.5cm}

%% file: sections/dataset.tex
In this work, we view the problem of floor plan generation from a text-based data structure perspective. Specifically, we wish to utilize the data structure to enable, through language input, a unified method that not only allows users to apply numerical constraint but also retains the ability to condition on graphs such as bubble diagrams, which have been the input in the traditional floorplan generation task.
As such, the choice of data representation for the floorplans is very important. In our work, we carefully design a JSON based data structure to be used as a new representation for the floorplans. As prior work all use datasets of floorplans in 2D image or 3D scenes, we create a new dataset by converting existing image and scene data into our data structure for the use of our task.


\subsection{Data Structure}
In this structure, we define crucial numerical data such as the number of rooms present, the total area and type of each room that appears in each floor plan. Each room within a floor plan is further defined with room specific fields. Importantly, each room's location is defined through a set of vertices that forms a polygon. This ensures that the language model cannot cheat by outputting vague locations, and instead it has to predict the exact coordinates of the vertices. The specific structure is presented in Table \ref{tab:datastructure}, and a full example of floorplan in this structure is presented in the Appendix \ref{appendix:gen_example}.

This structure also has several additional advantages: \textbf{1.} Using numerical value based gives us the ability to set numerical constraints in the input, and it allows for clear and explicit evaluation on how well the generation process adheres to these constraints. \textbf{2.} This structure is designed to have sufficient information to be transformed into higher level representations such as floor plan images, allowing seamless integration with traditional visualization techniques. \textbf{3.} The format is highly extensible, enabling the inclusion of additional information in the generation process, such as the placement and attributes of objects within the floorplan. This flexibility ensures that our approach can be further adapted to various task expansions in the future.
\begin{table}[h]
    \resizebox{\textwidth}{!}{\small\begin{tabular}{p{\linewidth}}
        \toprule
        \begin{algorithmic}
        \State \textbf{"room\_count"} \Comment{number of rooms}
        \State \textbf{"total\_area"} \Comment{total area in square units}
        \State \textbf{"room\_types" [ ]} \Comment{list of string of room types present}
        \State \textbf{"rooms" [ ]} \Comment{list of individual room dictionary containing room specifics}
        \State \quad \quad \textbf{"area"} \Comment{area of this room}
        \State \quad \quad \textbf{"floor\_polygon" [ ]} \Comment{list of vertices that defines this room's layout}
        \State \quad \quad \quad \textbf{"x"} \Comment{x coordinate of the vertex}
        \State \quad \quad \quad \textbf{"y"} \Comment{y coordinate of the vertex}
        \State \quad \quad \textbf{"is\_regular"} \Comment{flag that indicates if the room's shape is rectangular}
        \State \quad \quad \textbf{"height"} \Comment{the y-axis length of the rectangle bounding box enclosing the room}
        \State \quad \quad \textbf{"width"} \Comment{the x-axis length of the rectangle bounding box enclosing the room}
        \State \quad \quad \textbf{"id"} \Comment{unique id of this room}
        \State \quad \quad \textbf{"room\_type"} \Comment{the type of this room}
        \State \textbf{"edges" [ ]} \Comment{defines the connection between rooms for bubble gram generation}
        \end{algorithmic}\\
        \midrule
    \end{tabular}}
    \caption{Processed Floorplan Structure}
    \label{tab:datastructure}
\end{table}

\subsection{Datasets \& Preparation for Our New Task Formulation}
\label{sec:data}

\textbf{ProcTHOR-10k} \citep{procthor} is a dataset of 12,000 procedural generated, fully interactive 3D houses designed for research in Embodied AI. We clean and process the raw data from each house, focusing on the geometric properties of the rooms. We employ the shoelace formula to calculate areas and determined room dimensions based on their $x$ and $y$ coordinates. Additionally, we remove redundant points and apply rounding to the coordinates for consistency. Next, we categorize and count the types of rooms, and compute the total area for each house layout. The processed data is organized into the new JSON structure explained in Table \ref{tab:datastructure}.


\textbf{RPLAN} \citep{rplan} is a manually collected dataset of 80,788 real world floor plans of buildings in Asia.
Each floor plan in RPLAN is stored as a $256\times256\times4$ vector image. Channels 1 and 2 store interior and exterior boundary information; channel 3 contains room information where each pixel value denotes which room it belongs to; channel 4 has extra information to distinguish rooms with the same room type value in channel 3. To convert this 4 channel image into a JSON structure with well-defined room location, we first extract all pixel coordinates for all of the rooms. For each room, we locate the pixels that make up its perimeter. Then, by tracing the perimeter in one direction we are able to capture all of the vertices in an order that allows recreation of the room polygon. We take the room's type and deduce all the other fields listed in Table \ref{tab:datastructure} to complete the data structure. We convert 80,315 floorplans from RPLAN.

\subsection{Bubble Diagrams}
\label{subsec:bubble_diagram}
Traditionally, much of the prior work has assumed that at the beginning of the floor plan design process a bubble diagram is used to conceptualize the layout of the floor plan. As shown in Figure \ref{fig:generated_floorplans}, a bubble diagram is used to represent different room and their spacial relationship with one another as a guidance and constraint for the floor plan design process to follow. Formally, a bubble diagram is a graph $\mathcal{G} = (N,E)$ 
where each node $n_k$ in $N = \{n_j\}_1^R$ represents the $k^{th}$ room in the floor plan with $R$ rooms, and each edge $e_i = (n_p, n_q)$  in $E = \{e_j\}_1^M$ denotes a connection between room $p$ and $q$.

In this work, we enable the use bubble diagram as an additional conditioning in our floor plan generation process. To obtain the bubble diagram we check pair-wise proximity of the rooms in each floor plan. If the Manhattan distance between the rooms' boundaries are within a threshold, they are counted as connected. This threshold is set to 8 pixels for the RPLAN dataset in line with prior work, and to 2 pixels for ProcTHOR to account for the length unit differences between the two datasets. This adjacency information is stored in the "edges" field in our data representation. During training, we use $\mathcal{G}$ as conditioning for our floor plan generation.

\begin{figure}[h]
    \centering
    \includegraphics[width=\linewidth]{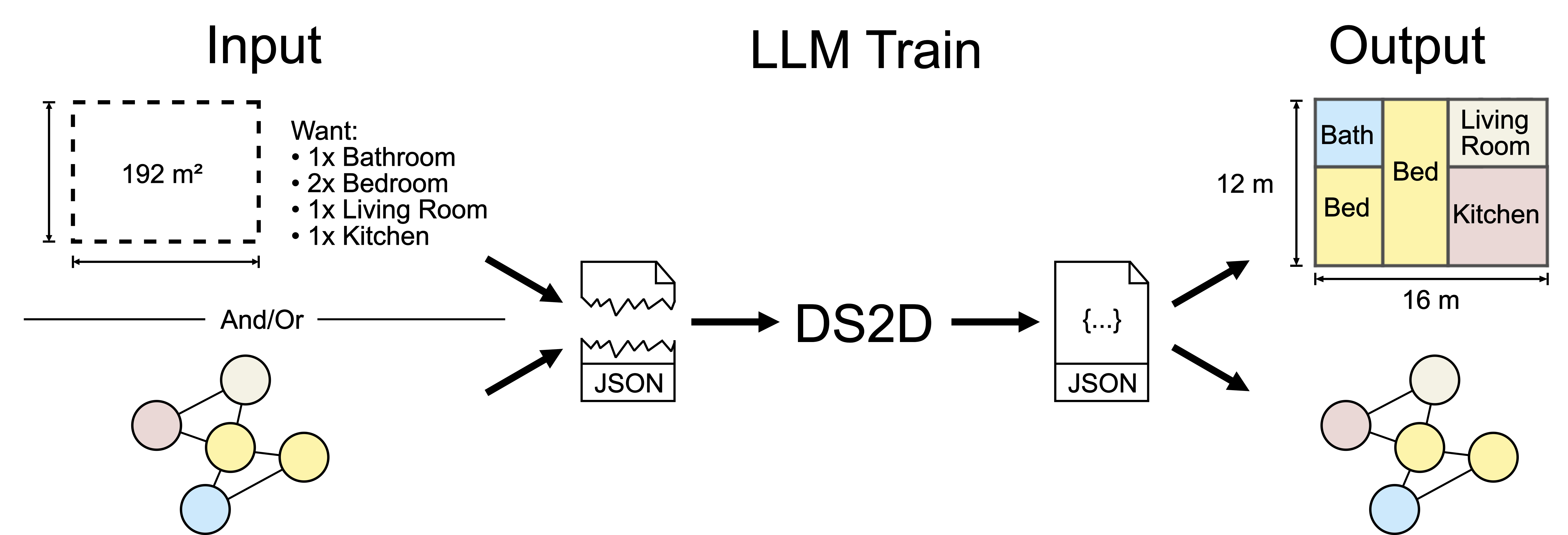}
    \caption{Model Pipeline Overview}
    \label{fig:training_overview}
\end{figure}

%% file: sections/llm.tex
We train an LLM by prompting it with structured numerical constraints and or graph constraints (Bubble Diagram) and asking it to predict the converted floor plan, as depicted in Figure \ref{fig:training_overview}.

\textbf{Numerical Constraints} $\mathcal{C} = \{c_j\}_1^T$ are a set of $T$ conditions that users may impose on the final floor plan. In the task of floor plan design and generation, these can include, but are not limited to, total square footage of the entire floor, the number and the types of rooms present, the size of each room. 
These conditions are mostly inherently numerical, and thus we can take advantage by directly using it in its data structure form. 
For example, the constraint of having a total square footage of 990 can be transformed into \verb|{"total_area": 990}|. 
This allows a direct match between the input prompt and output JSON string, which may help the model understand the structure and its relationship better.

\textbf{Bubble Diagrams} are represented differently as conditioning to our model. We chose to pass it in as tuples of connecting rooms. For an edge $e_i = (n_p, n_q)$ connecting nodes $p$ and $q$, it is formatted as \verb|(room_p, room_q)|. Due to the possible existence of multiple rooms of the same type, which can be clearly distinguished in a graph but not easily in text, we decide to represent \verb|room_p| and \verb|room_q| by their room IDs in addition to their room types. 



We process the constraint set and the bubble diagram into "constraint string" and "adjacency string". They are then combined with an instruction phrase that is used throughout the training process to create the final prompt. An example of the full prompt is shown in Appendix \ref{appendix:prompt}

%% file: sections/metrics.tex
\begin{table}[h]
    \centering
    \small\begin{tabular}{p{0.15\linewidth}p{0.19\linewidth}p{0.55\linewidth}}
            \toprule
             \textbf{Metric Type} & \textbf{Metric}  &  \textbf{Explanation}  \\
            \midrule
             \multirow{11}{2cm}{Self Consistency (SC)}        & $\dag$ Total Area & percentage difference between stated total area vs. the sum of all rooms' state area\\ \cmidrule{2-3}
              &  $\ddag$ \makecell[tl]{Polygon Area \\ (P. Area)} & percentage difference between a room's stated area vs. area calculated from polygon \\ \cmidrule{2-3}
              & $\sharp$ \makecell[tl]{Overlap} &  boolean check for existence of polygon overlap \\\cmidrule{2-3}
              & $\flat$ Room ID (ID)  &  boolean check for duplicate room id \\ \cmidrule{2-3}
            & $\natural$ \makecell[tl]{Room Count \\ (R. Count)} &  boolean check to see if "room\_count" field number is equal to the number of rooms in the floor plan.  \\ \midrule
          \multirow{4}{2cm}{Both SC ~~ ~~~and PC} & $\triangleleft$ \makecell[tl]{Room Height \\(R. H)} &  percentage difference between height obtained from polygon vs. stated height in generation (SC) or requested height (PC).\\ \cmidrule{2-3}
              & $\triangleright$ \makecell[tl]{Room Width\\ (R. W)} &   percentage difference between width obtained from polygon vs. stated width in generation (SC) or requested width (PC).\\
            \midrule
             \multirow{15}{2cm}{Prompt Consistency (PC)}   &  $\bullet$ \makecell[tl]{Total Area \\} & percentage difference between the sum of room polygon areas vs. requested total area \\ \cmidrule{2-3}
             &  $\clubsuit$ \makecell[tl]{Num. Room \\ (Num. R.)} & percentage difference between number of rooms in the floorplan vs. requested number of rooms in the prompt. \\ \cmidrule{2-3}
            & $\spadesuit$ \makecell[tl]{Room ID\\ (ID)} &  precision and recall of room ids in the floorplan with respect to the ids present in the prompt\\ \cmidrule{2-3}
              & $\heartsuit$ \makecell[tl]{ Room Area \\ (R. Area)} &   percentage difference between a room's polygon area vs. requested room area in the prompt \\ \cmidrule{2-3}
           & $\diamondsuit$ \makecell[tl]{Room Type \\(Type)} &  precision and recall of room types in the floorplan with respect to the room types present in the prompt \\ \cmidrule{2-3}
              & $\circ$ \makecell[tl]{ID-Type Match \\(IDvsType)} & percentage difference between the number of rooms with correctly matching type and id in the floor plan vs. the total number of rooms present in the prompt\\
            \midrule
            Bubble Graph   &  Compatibility &   graph edit distance (GED) between the input bubble diagram and one extracted from the output floor plan.\\
            \bottomrule
    \end{tabular}
    \caption{Metrics and Associated Acronyms Used to Evaluate Generation Quality}
    \label{tab:metrics}
\end{table}
To assess floorplan quality, we focus on the numerical consistency of the generation. Specifically, we design two groups of metrics to evaluate what we call self-consistency and prompt-consistency:
\begin{itemize}[leftmargin=10pt]
\item \textbf{Self Consistency} metrics measures how numbers agree with each other in the generated floor plan. For instance, this includes a metric to check if the area defined by the polygon vertices is the same as the area number presented in the "room\_area" field. 
\item \textbf{Prompt Consistency} metrics evaluate how consistent the generation is to the constraints used in the prompt. An example of this type of metrics is one that measures if the generated number of rooms adds up to the number of room requested in the prompt.
\end{itemize}

In addition to the two main groups of metrics, we also incorporate \textbf{Compatibility} metrics which is used in past work \citep{shabani2022housediffusion,nauata2020housegan,nauata2021housegan,johnson2018image, ashual2019specifying} to measure similarity between the input bubble diagram and the output floor plan. 
It is by definition the graph edit distance \citep{abu2015exact} between the input bubble diagram and the output diagram extracted from the output JSON. The extraction method is the same one used to generate the bubble diagrams in the first place as described in Section \ref{subsec:bubble_diagram}. The full array of metrics and their explanations are presented in Table \ref{tab:metrics}.

%% file: sections/experiments.tex
All of our experiments are ran on the LLaMA3-8B-Instruct variant of the LLaMA model family \citep{llama}. We train our models by running 8-bit quantization along with LoRA \citep{hu2021lora}. Each of our experiments are also run on single RTX8000 GPU, they take less than a day to complete. The exact training parameters and data split details can be found in Appendix \ref{appendix:train_parameters}

\begin{table}[h]
    \centering
    \small\begin{tabular}{c c c c c c c p{0.7cm} p{0.7cm} p{0.7cm} p{0.7cm}}
            \toprule
                &   \multicolumn{6}{c}{ProcTHOR Models} & \multicolumn{4}{c}{RPLAN Models}  \\
            & F & M & PM & F+BD & M+BD & PM+BD & ~~ 5-R & ~~6-R & ~~ 7-R & ~~8-R \\
            \cmidrule(lr){2-7} \cmidrule(lr){8-11}
        Bubble Diagram     &  - & - & - & \checkmark &\checkmark &\checkmark &  ~~~~\checkmark & ~~~~\checkmark & ~~~~\checkmark & ~~~~\checkmark \\
        Masking   & - &  random  & preset    & - &  random & preset & random & random &  random &  random  \\
            \bottomrule
    \end{tabular}
    \caption{Summary of model variants}
    \label{tab:variants}
\end{table}
\vspace{-0.5cm}
\subsection{Model Variants}\label{subsec:model_variants}
We train six variants of our model on the converted ProcTHOR dataset and four on converted RPLAN dataset. 
Each of the four models on RPLAN -- \textbf{5-R}, \textbf{6-R}, \textbf{7-R}, \textbf{8-R} -- is trained without floorplan data of a certain room count.
5-R is trained only on floorplans with 6, 7, and 8 rooms, etc. This follows prior work \citep{nauata2020housegan,nauata2021housegan,shabani2023housediffusion}.
On the other hand, the six models trained on ProcTHOR are divided two main variants: bubble diagram enabled model, and numerical constraint only model. In the \textbf{bubble diagram enabled model (BD)}, both the constraint string and the adjacency string described in Section \ref{sec:llm} are used as part of the input. In constrast, in the \textbf{numerical constraint only model}, only the constraint string is used. We further train three sub-variants for each of the two main variants to investigate how robust the models are to missing information:
\begin{itemize}[leftmargin=10pt]
\item \textbf{Full-Prompt (F)} model takes advantage of every attribute available in the constraint set. During training, every constraint attribute is part of the constraint string.
\item \textbf{Mask (M)} model uses a subset of constraint set by applying a 50\% masking on every single possible constraint. In the case of \texttt{rooms} which is an attribute containing a list, the masking is applied independently to individual item in the list. (\texttt{rooms} is only dropped out if the entire list becomes empty.) As a safety measure, we always keep at least 1 constraint in the set.
\item \textbf{Preset Mask (PM)} randomly selects one of four preset constraint sets with varying degree of missing information. The idea is to not just have IID random masking, but a hierarchy of attributes from general to specific which are detailed in Table \ref{tab:presets} in Appendix \ref{appendix:preset_masking}.
\end{itemize}
The summary of our six variants and their abbreviations are shown in Table \ref{tab:variants}

\subsection{Generation Prompts}
\label{subsec:prompt_strat}
To test the model, we run four different generation prompts with varying amount of constraints in the prompt. This simulates the real world problem of floorplan design, where designer often receives only partial criteria. For instance, the user might only ask the living room and bedroom to be a certain size, and leaves the designer with the freedom to imagine layouts with varying sizes for the other rooms. The four generation prompts are listed in order of decreasing amount of constraints:
\begin{itemize}[leftmargin=10pt]
\item \textbf{Specific (S)} is one where we use all the possible constraints
\item \textbf{All Room Area (AR)} is one where we pass in the area of all the rooms (total area can be inferred from this information).
\item \textbf{Partial Room Area (PR)} is similar to Total Area. In addition to the total area of the floor plan, the area of some of the rooms are also passed as conditions in the constraint string. 
\item \textbf{Total Area (TA)} is one where we only use the total area of the floor plan as constraint.
\end{itemize}

For all of the Bubble Diagram model variants, Bubble Diagrams are used in each of these types of prompts.

%% file: sections/Results.tex
\begin{figure}[h]
    \centering
    \includegraphics[width=0.9\linewidth]{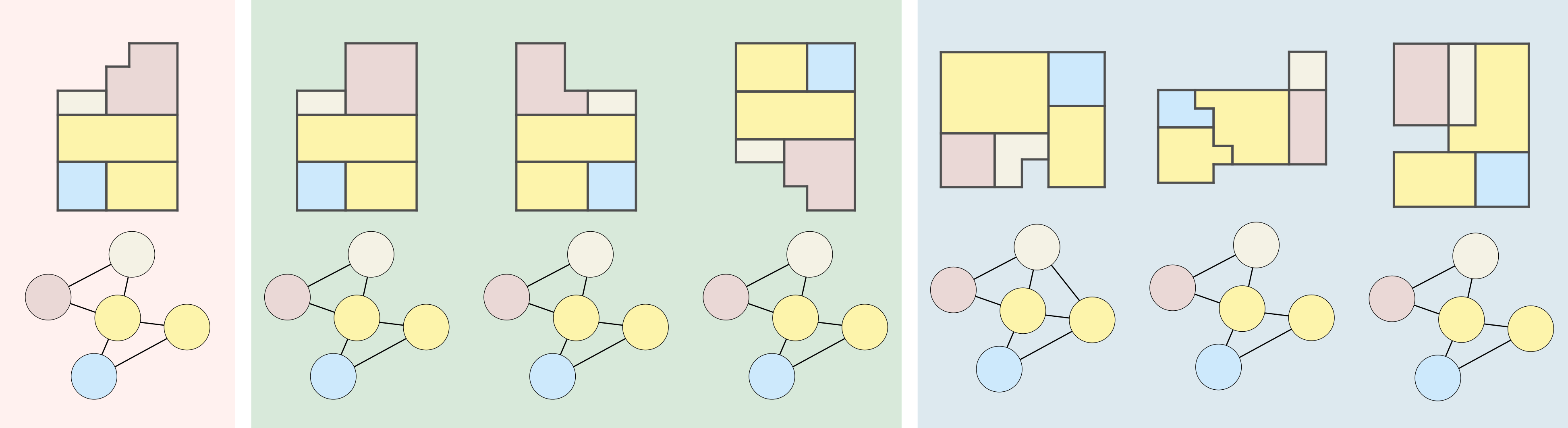}\\
    \includegraphics[width=0.9\linewidth]{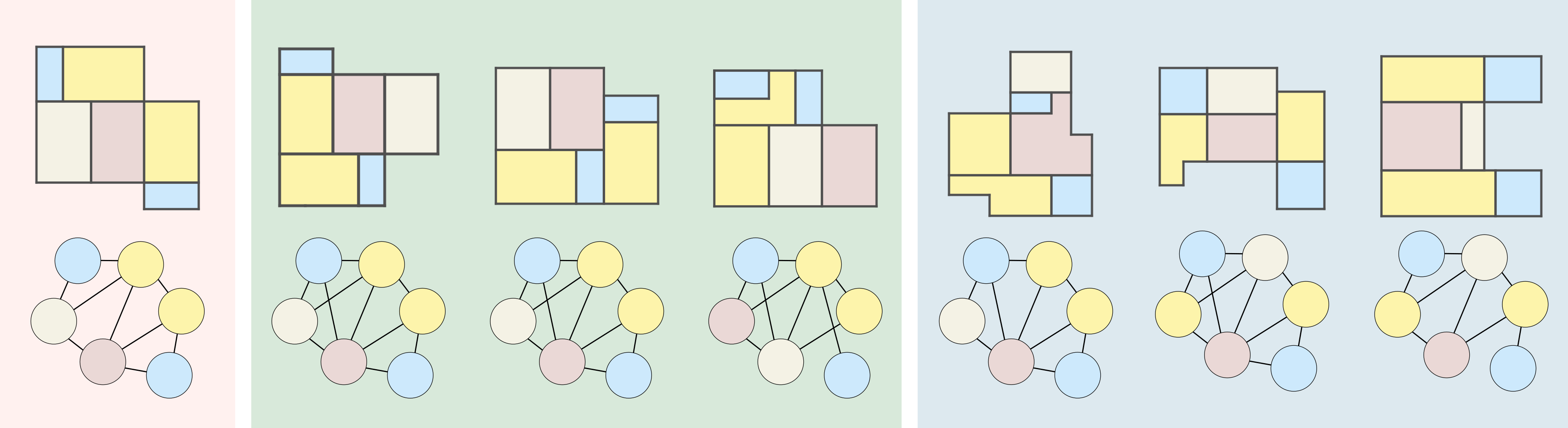}\\ 
    \vspace{0.1cm}
    \includegraphics[width=0.5\linewidth]{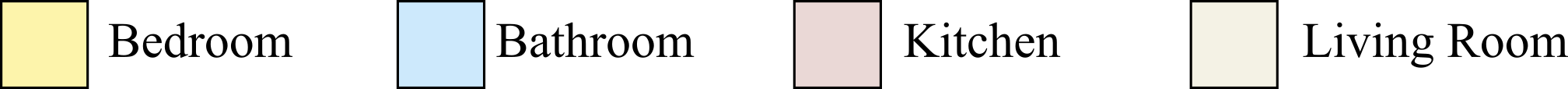}
    \caption{Example generations and extracted bubble diagrams. Pink background: ground truth floor plan and its associated bubble diagram. Green background: from F+BD model with Specific prompts. Blue background: from F+BD model with Total Area prompts.}
    \label{fig:generated_floorplans}
\end{figure}

\begin{table}[h] 
    \centering
    \small\begin{tabularx}{\textwidth}{lccccp{1.5cm}cc}
    \toprule
       &\multicolumn{7}{c}{\textbf{Self-consistency Benchmark}}\\
     
    & P. Area$^\ddag$  & Total Area$^\dag$ & P. Overlap$^\sharp$ & R. Count$^\natural$  & ~~~~~~~~ID$^\flat$  & R. H$^\triangleleft$ & R. W$^\triangleright$ \\
    \cmidrule{2-8}
       F\textsubscript{S}    & 0.94$\pm 0.07$   & 1.00$\pm 0.03$ & 0.16$\pm 0.37$  & 1.00 & ~~~~~~1.00   & 0.99$\pm 0.02$  & 0.99$\pm 0.02$  \\
         M\textsubscript{AR}  & 0.90$\pm 0.07$  & 0.99$\pm 0.03$  & 0.12$\pm 0.33$  & 1.00  & ~~~~~~1.00  & 0.99$\pm 0.02$  & 0.99$\pm 0.01$  \\
        PM\textsubscript{PR}  & 0.92$\pm 0.06$    & 0.97$\pm 0.07$  & 0.15$\pm 0.35$  & 0.96  & ~~~~~~1.00  & 1.00$\pm 0.01$  & 1.00$\pm 0.01$\\
        PM\textsubscript{TA}   & 0.93$\pm 0.05$  & 0.99$\pm 0.04$  & 0.14$\pm 0.35$  & 0.95  & ~~~~~~1.00  & 1.00$\pm 0.01$  & 1.00$\pm 0.01$ \\
         \midrule
         5-R\textsubscript{AR} & 0.90$\pm 0.16$   & 0.98$\pm 0.05$  & 0.37$\pm 0.48$ & 0.97  & ~~~~~~1.00   & 1.00$\pm 0.01$  & 0.98$\pm 0.02$\\
         6-R\textsubscript{S}    & 0.93$\pm 0.05$ & 0.98$\pm 0.06$  & 0.36$\pm 0.48$ & 0.99   & ~~~~~~1.00 & 0.99$\pm 0.01$  & 0.97$\pm 0.02$\\
         7-R\textsubscript{S}    & 0.93$\pm 0.04$  & 0.98$\pm 0.05$ & 0.42$\pm 0.49$  & 0.97   & ~~~~~~1.00 & 0.99$\pm 0.01$  & 0.98$\pm 0.02$  \\
         8-R\textsubscript{TA}    & 0.82$\pm 3.50$ & 0.97$\pm 0.07$ & 0.53$\pm 0.50$  & 0.89  & ~~~~~~1.00 & 0.99$\pm 0.01$   & 0.98$\pm 0.02$  \\
         \bottomrule
    \end{tabularx}
    \small\begin{tabularx}{\textwidth}{lcccccccc}
    \toprule
     & \multicolumn{7}{c}{\textbf{Prompt-consistency Benchmark}}\\
     
     & Num. R$^\clubsuit$ & Total Area$^\bullet$ &  R. Area$^\heartsuit$ & ID$^\spadesuit$ & IDvsType$^\circ$  &  R. H$^\triangleleft$   &  R. W$^\triangleright$ \\
    \cmidrule{2-8}
         F\textsubscript{S}    & 1.00$\pm 0.00$  & 0.95$\pm 0.07$  & 0.94$\pm 0.07$   & 1.00$\pm 0.03$  & 1.00$\pm 0.00$  & 0.99$\pm 0.02$   & 0.99$\pm 0.02$\\
     M\textsubscript{AR} & 1.00$\pm 0.00$  & -  & 0.90$\pm 0.07$ & 1.00$\pm 0.00$  & 1.00$\pm 0.00$  & - & -  \\
    PM\textsubscript{PR}  & 1.00$\pm 0.00$  & 0.93$\pm 0.07$  & 0.91$\pm 0.11$   & 1.00$\pm 0.03$ & 1.00$\pm 0.00$ & -  & -\\
         PM\textsubscript{TA}  & 1.00$\pm 0.00$  & 0.95$\pm 0.06$   & -  & -  & -  & -  & - \\
     \midrule
         5-R\textsubscript{AR}    & -  & -  & 0.92$\pm 0.04$ & 1.00$\pm 0.01$  & 1.00$\pm 0.00$  & -  & - \\
         6-R\textsubscript{S}   & 1.00$\pm 0.00$  & 0.94$\pm 0.07$  & 0.94$\pm 0.06$    & 1.00$\pm 0.05$  & 1.00$\pm 0.00$ & 0.92$\pm 0.23$  & 0.94$\pm 0.14$ \\
        7-R\textsubscript{S}   & 1.00$\pm 0.00$  & 0.94$\pm 0.06$  & 0.94$\pm 0.06$ & 1.00$\pm 0.02$  & 1.00$\pm 0.00$  & 0.93$\pm 0.17$    & 0.94$\pm 0.29$ \\
        8-R\textsubscript{TA} & -  & 0.93$\pm 0.07$  & -  & -  & -  & -  & -  \\
    \bottomrule
    \end{tabularx}
    \caption{Our self- and prompt-consistency Benchmark results. Subscripts under model variants are generation prompts.}
    \label{tab:main_results}
\end{table}

We test floor plan generation quality on ProcTHOR-trained six model variants with the four different generation prompts with self and prompt consistency metrics. Results from our best model variants are listed in Table \ref{tab:main_results}, and all results on all 48 sets of experiments are listed in Appendix \ref{appendix:procthor_metric_array}.

We perform the same evaluation on the four RPLAN-trained model variants. Because each model variant trained on RPLAN uses a slightly different set of data (following prior work), we show results for each model variant in Table \ref{tab:main_results}. Results for the full 24 experiments are shown in Appendix \ref{appendix:rplan_consistency_metrics}.

\paragraph{Analysis} According to our metrics, perhaps surprisingly, our LLM models demonstrate a high level of competence in generating floorplans that are largely mathematically consistent, and are also consistent to input numerical constraint. As seen in Figure \ref{fig:generated_floorplans}, when given full specifications, the generated rooms largely resembles ground truth in sizes, width and height. This quality is accurately reflected by our Total Area, R.H and R.W metrics. On the other hand, our metric suggests that generations can produce overlapping room layouts. An example of this is found in Figure \ref{fig:generated_floorplans}, in the second row's green background's right-most floorplan: the top-left bathroom is overlapping on top of the bedroom. This problem is especially apparent in the RPLAN-trained models, and is accurately reflected by the P. Overlap metrics in Table \ref{tab:main_results}. This limitation can perhaps be tackled in future work.

To compare with prior work that focus on using bubble diagrams as input on the RPLAN dataset, we prompt our RPLAN-trained models with bubble diagram. For evaluation we compute the standard compatibility metric obtained from graph edit distance calculation between the input bubble diagrams and the ones extracted from generations. Results are shown in Table \ref{tab:rplan_ged}. 
We also evaluate compatibility metrics on the ProcTHOR-trained BD model variants. The results are shown in Table \ref{tab:procthor_ged_best}. 

\begin{table}[h]
\centering
\small\begin{tabular}{*5c}
\toprule
 &  \multicolumn{4}{c}{Number of Rooms in Floorplan}\\
\cmidrule{2-5}
Model    & 5   & 6    & 7   & 8\\
\midrule
\citet{ashual2019specifying} & 7.5 $\pm$ 0.0 & 9.2 $\pm$ 0.0  & 10.0 $\pm$ 0.0  & 11.8 $\pm$ 0.0  \\
\citet{johnson2018image} & 7.7 $\pm$ 0.0  & 6.5 $\pm$ 0.0  & 10.2 $\pm$ 0.0  & 11.3 $\pm$ 0.1  \\
House-GAN \citep{nauata2020house}  &  2.5 $\pm$ 0.1  & 2.4 $\pm$ 0.1   & 3.2 $\pm$ 0.0   & 5.3 $\pm$ 0.0 \\
House-GAN++ \citep{nauata2021house}   &  1.9 $\pm$ 0.3  & 2.2 $\pm$ 0.3    & 2.4 $\pm$ 0.3   & 3.9 $\pm$ 0.5 \\
HouseDiffusion \citep{shabani2023housediffusion}   &  1.5 $\pm$ 0.0  &  1.2 $\pm$ 0.0    & 1.7 $\pm$ 0.0   & \textbf{2.5} $\pm$ 0.0 \\
\midrule
Our BD Model w/ BD prompt    & \textbf{0.46} $\pm$ 0.73 & \textbf{0.79} $\pm$ 0.98  & \textbf{1.27} $\pm$ 1.28 & \textbf{2.50 }$\pm$ 1.97\\
\bottomrule
\end{tabular}
    \caption{Comparison of Compatibility$\downarrow$ score (Graph Edit Distance) on the RPLAN dataset.}
    \label{tab:rplan_ged}
\end{table}
\paragraph{Analysis} 
Overall, our RPLAN-trained model generates results that are competitive with the state-of-the-art image based algorithms, even though our focus here has been on numerical accuracy. One caveat is that the bad generations are very wrong, causing our method to have a much higher error rate compared to image-based methods.
Table \ref{tab:procthor_ged_best} suggests our model has a harder time following the input bubble diagram as the number of room increases, or when given room area information. This is reflected in Figure \ref{fig:generated_floorplans} where the second row's bubble diagrams show more variations. This limitation might be tackled by using a better data structure to represent bubble diagrams, and perhaps also to be included as part of the floorplan data representation. This way, the bubble diagram will simply be part of the constraint set, and this may help the model understand the input better. This can be explored in future work in the same task setting. Additionally, in the future, our data can be expanded to account for more floorplan attributes such as doors and walls and our benchmark can include additional metrics to account for more obscure numerical checks.


\begin{table}[h]
\centering
\small\begin{tabular}{ccccccc}
\toprule
Model & Generation & \multicolumn{5}{c}{Number of Rooms}\\
\cmidrule{3-7}
Variant &  Prompt   & 3   & 4  & 5 & 6 & 7  \\
\midrule
F+BD        & Specific   & 0.15 $\pm$0.41 & 0.33 $\pm$0.58 & 1.35 $\pm$1.34 & 2.24 $\pm$1.76 & 4.23 $\pm$1.89 \\
            & All Room Area   & 5.91 $\pm$0.29 & 9.75 $\pm$0.54 & 13.79 $\pm$1.15 & 18.09 $\pm$1.71 & 22.35 $\pm$2.24  \\
            & Partial Room Area &  3.80 $\pm$1.20  & 5.73 $\pm$2.03  & 8.19 $\pm$2.72 & 10.15 $\pm$2.88  & 11.42 $\pm$4.52 \\
            & Total Area    & 0.19 $\pm$0.40 & 0.62 $\pm$0.78 & 3.48 $\pm$1.83 & 3.41 $\pm$1.50 & 6.23 $\pm$2.20   \\
\midrule
M+BD &Specific   &  0.16 $\pm$0.45 &  0.47 $\pm$0.67 &  2.06 $\pm$1.33 &  2.85 $\pm$1.79 &  5.23 $\pm$2.14 \\
        &All Room Area   &5.91 $\pm$0.29 & 9.62 $\pm$0.56 & 12.90 $\pm$1.27 & 17.48 $\pm$1.82 & 22.13 $\pm$1.94  \\
        &Partial Room Area    & 3.85 $\pm$1.11 & 6.35 $\pm$1.95 & 8.44 $\pm$2.48 & 11.56 $\pm$3.42 & 14.13 $\pm$3.90   \\
        &Total Area   & 0.20 $\pm$0.40 & 0.54 $\pm$0.73 & 2.40 $\pm$1.03 & 3.09 $\pm$2.07 & 6.26 $\pm$2.03  \\
\midrule
PM+BD &Specific  & 0.15 $\pm$0.39 & 0.58 $\pm$0.96 & 2.42 $\pm$1.33 & 2.97 $\pm$1.62 & 5.97 $\pm$1.25 \\
        &All Room Area   & 5.88 $\pm$0.33 & 9.77 $\pm$0.49 & 13.17 $\pm$1.23 & 18.56 $\pm$1.58 & 21.97 $\pm$2.36   \\
        &Partial Room Area   & 3.87 $\pm$1.13 & 6.42 $\pm$1.97 & 8.99 $\pm$2.67 & 12.06 $\pm$3.05 & 13.68 $\pm$3.71    \\
        &Total Area   &  0.34 $\pm$0.93 & 0.46 $\pm$0.84 & 2.08 $\pm$1.29 & 2.65 $\pm$1.87 & 5.84 $\pm$1.61   \\
\bottomrule
\end{tabular}
    \caption{Compatibility$\downarrow$ score on ProcTHOR-trained models.}
    \label{tab:procthor_ged_best}
\end{table}

%% file: sections/conclusion.tex

%% file: sections/appendix.tex
\appendix

\section{Dataset License and Privacy}
\label{appendix:license}


The \textbf{ProcTHOR-10k} dataset \citep{procthor} is procedurally generated under Apache License 2.0 and distributed with the Allen Institute for AI's prior package. 

The \textbf{RPLAN} dataset \cite{rplan} is created with anonymized floorplan data that eliminates user and privacy information. Furthermore, the original paper states that the floorplans used are ensured to have no copyright issues. Besides, researchers are not permitted to redistribute the downloaded data, in whole or in part, through other media. 

Due to redistribution restrictions of the \textbf{RPLAN}, we cannot directly share our new dataset. However, after downloading the \textbf{RPLAN} from the author's source, the code provided in the supplementary materials can be used to create our new dataset.

\section{Possible Negative Impacts}
\label{appendix:impact}

\paragraph{Job Displacement}
The automation of floorplan generation through advanced technology threatens job displacement for professionals in architecture, real estate, and related fields. As these tools become more sophisticated, human expertise may be undervalued or rendered obsolete.
\paragraph{Mitigation} To address this issue, it is important to promote technology as a means to augment human capabilities rather than replace them. By providing training programs, professionals can adapt to new tools and workflows, ensuring they remain relevant and can leverage technology to enhance their work rather than be supplanted by it.

\paragraph{Quality and Safety}
There is a significant risk that generated floorplans might not meet established safety standards or quality expectations, leading to potential hazards in construction or living conditions. Poorly designed floorplans could result in unsafe buildings and legal liabilities and reduce occupants' overall quality of life. 
\paragraph{Mitigation} To mitigate these risks, it is essential to incorporate rigorous safety and quality checks into the floorplan generation process. Collaborating with experts in architecture and engineering can help ensure that generated floorplans adhere to all relevant standards and regulations, thus maintaining the integrity and safety of the built environment.

\section{Prompt Structure}
\label{appendix:prompt}
\begin{verbatim}
    <|start_header_id|>system<|end_header_id|>
    you are to generate a floor plan in a JSON structure where each room 
    is defined by polygon vertices, make sure to not overlap the polygons. 
    you have to satisfy the adjacency constraints given as pairs of 
    neighboring rooms; two connecting rooms, room1 and room2, are presented 
    as (room1_type/"room1_id", room2_type/"room2_id"). you have to also 
    match the specifications passed by the user in a JSON structure when 
    they exist. when room area and total area requirements exist, make sure 
    the polygon areas add up to the required number.
    <|eot_id|><|start_header_id|>user<|end_header_id|>
    adjacency constraints: (Bedroom/"room|4", Bathroom/"room|5"),
    (Bedroom/"room|4", Kitchen/"room|6"), (Bedroom/"room|4",
    LivingRoom/"room|7"),(Bathroom/"room|5", Kitchen/"room|6"), 
    (Bathroom/"room|5", LivingRoom/"room|7"),
    (Kitchen/"room|6", LivingRoom/"room|7"). specifications: 
    {\'room_count\': 4, \'total_area\': 146.8, \'rooms\': [
    {\'area\': 41.3, \'id\': \'room|4\', \'room_type\': \'Bedroom\'},
    {\'area\': 27.5, \'id\': \'room|7\', \'room_type\': \'LivingRoom\'}]}
    <|eot_id|><|start_header_id|>assistant<|end_header_id|>
\end{verbatim}

\section{Training Details}
\label{appendix:train_parameters}
\subsection{Dataset Details}
For floorplan data converted from \textbf{RPLAN} dataset, we follow the original data split. RPLAN dataset is divided into 5 parts according to the number of rooms present in the floorplan. Specifically, each part respectively contains floorplans with 4, 5, 6, 7, 8 rooms. Each part is further divided into a training, validation, and a test split. Following prior work, we use only training data with 5, 6, 7, 8 rooms. 4 different models are trained, each has never seen one of the 4 part -- when training the model for 5 room floorplan generation, training is done with training splits of 6, 7, 8 room floorplans etc. During test time, the model is asked to generate only 4 room floorplans even though it has never seen a 4 room floorplan in the training set. This is in line with prior work that uses the RPLAN dataset.

Forfloorplan data converted from \textbf{ProcTHOR} dataset, the validation and test split is randomly chosen, with each of validation and test split being 10\% of the training split. The training split is used to train the model; the validation split used to perform early stopping; and the test split is used to generate prompts for generation and evaluation.

\subsection{Hyper-parameters}
We follow the suggest training parameters from Llama-recipes (https://github.com/meta-llama/llama-recipes/tree/main), and we do not perform any hyper-parameter search.

\textbf{Random Seed} used for generation is chosen to be 12345 to select 100 (sampling generation) or 1000 (greedy generation) random test set data for generation. This seed is simply chosen and never changed or compared with any other random seed.

For \textbf{Sampling} generation, we generate 20 output floorplans for each prompt using nucleus sampling. We choose $p=0.8$ for sampling as it is a common value for this parameter.

For \textbf{Greedy} generation, we use the default parameter provided by HuggingFace \citet{huggingface_gen}.

\section{Specific Attributes Sets in Preset Masking}
\label{appendix:preset_masking}
Here we present the attributes used in the preset masking variant of our model.
The four presets have varying amount of attributes from general to specific.
\begin{table}[h!]
    \centering
    \begin{tabular}{cc}
            \toprule
              Preset  &  Constraint Attributes  \\
            \midrule
            1 &    room\_count, room\_types, total\_area \\
            2 &    room\_count, room\_types,  total\_area, \textbf{partial list of (}rooms[id], rooms[type], rooms[area]\textbf{)}\\
            3 &    room\_count,  room\_types,  \textbf{partial list of (}rooms[id], rooms[type], rooms[area]\textbf{)}\\
            4 &   room\_count,  room\_types, \textbf{full list of (}rooms[id], rooms[type], rooms[area]\textbf{)}\\
            
            \bottomrule
    \end{tabular}
    \caption{Different presets of constraint set. Partial list represents information for only some of the rooms: dropout is applied on the individual room in the list of all rooms}
    \label{tab:presets}
\end{table}

\section{Self and Prompt Consistency Metrics on ProcTHOR}
\label{appendix:procthor_metric_array}
In total, we have 6 model variants trained on converted ProcTHOR data. 3 of them are trained with bubble diagram input and 3 without. For each of the 6 model variants, we run 4 different sets of generations with different prompts as explained in Section \ref{subsec:prompt_strat}; and for each of the generation prompt, we run both greedy and sampling algorithm to obtain 2 sets of different floorplan generations.

Therefore, in total, we run 48 sets of experiments and evaluations. We present the full evaluation using our designed metrics here, organized into 4 different sections according to the generation prompt. Each section has 2 main tables, one for self consistency metrics, and one for prompt consistency metrics.

\newpage
\subsection{Generation Prompt: Specific}
\label{appendix:gen_procthor_specific}
\begin{table}[h]
    \centering
    \small\begin{tabular}{lccccccc}
    \toprule
       &\multicolumn{7}{c}{\makecell[tc]{ Greedy Generation \\\bf Self Consistency Metrics}}\\
{}  &  P. Area  &  Overlap  &  R. Count  &  R. H  &   ID  &  R. W  &  Total Area  \\
    \cmidrule{2-8}
                F    & 0.94$\pm 0.07$  & 0.16$\pm 0.37$  & 1.00  & 0.99$\pm 0.02$  & 1.00  & 0.99$\pm 0.02$  & 1.00$\pm 0.03$ \\
  F+BD    & 0.95$\pm 0.06$  & 0.22$\pm 0.41$  & 1.00  & 0.99$\pm 0.02$  & 1.00  & 0.99$\pm 0.04$  & 1.00$\pm 0.00$ \\
     M    & 0.93$\pm 0.07$  & 0.18$\pm 0.39$  & 1.00  & 0.99$\pm 0.04$  & 1.00  & 0.98$\pm 0.06$  & 1.00$\pm 0.00$ \\
  M+BD    & 0.93$\pm 0.07$  & 0.19$\pm 0.39$  & 1.00  & 0.99$\pm 0.04$  & 1.00  & 0.99$\pm 0.03$  & 1.00$\pm 0.03$ \\
    PM    & 0.92$\pm 0.06$  & 0.12$\pm 0.32$  & 0.98  & 0.98$\pm 0.07$  & 1.00  & 1.00$\pm 0.01$  & 1.00$\pm 0.02$ \\
 PM+BD    & 0.90$\pm 0.09$  & 0.17$\pm 0.37$  & 1.00  & 1.00$\pm 0.03$  & 1.00  & 1.00$\pm 0.01$  & 1.00$\pm 0.00$ \\
    \midrule
     &\multicolumn{7}{c}{\makecell[tc]{Sampling Generation \\\bf Self Consistency Metrics}}\\
     {}  &  P. Area  &  Overlap  &  R. Count  &  R. H  &  RID  &  R. W  &  Total Area  \\
    \cmidrule{2-8}
            F    & 0.95$\pm 0.06$  & 0.17$\pm 0.38$  & 1.00  & 0.99$\pm 0.02$  & 1.00  & 0.99$\pm 0.02$  & 1.00$\pm 0.00$ \\
   F+BD    & 0.95$\pm 0.06$  & 0.22$\pm 0.41$  & 1.00  & 0.99$\pm 0.03$  & 1.00  & 0.99$\pm 0.03$  & 1.00$\pm 0.00$ \\
      M    & 0.93$\pm 0.08$  & 0.16$\pm 0.37$  & 1.00  & 0.99$\pm 0.04$  & 1.00  & 0.98$\pm 0.07$  & 1.00$\pm 0.00$ \\
   M+BD    & 0.93$\pm 0.07$  & 0.17$\pm 0.38$  & 1.00  & 0.99$\pm 0.04$  & 1.00  & 0.99$\pm 0.04$  & 1.00$\pm 0.02$ \\
     PM    & 0.91$\pm 0.07$  & 0.14$\pm 0.35$  & 0.97  & 0.96$\pm 0.09$  & 1.00  & 1.00$\pm 0.02$  & 1.00$\pm 0.02$ \\
  PM+BD    & 0.90$\pm 0.08$  & 0.20$\pm 0.40$  & 1.00  & 0.99$\pm 0.06$  & 1.00  & 1.00$\pm 0.00$  & 1.00$\pm 0.00$ \\
    \bottomrule
\end{tabular}
    \caption{Evaluation of \textbf{self consistency metrics} on generated results from \textbf{Specific Prompt} (Section \ref{subsec:prompt_strat}). Top half are results on single greedy generations from 1000 prompts from 1000 different test set floorplans; bottom half are results on sampled generations on 100 different prompt with 20 generations from each prompt.}
    \label{tab:detailed_eval}
\end{table}

\begin{table}[h]
    \centering
    \resizebox{\textwidth}{!}{\small\begin{tabular}{lcccccccc}
    \toprule
       &\multicolumn{8}{c}{\makecell[tc]{ Greedy Generation \\\bf Prompt Consistency Metrics}}\\
{}   & Num. R. & Total Area &  R. Area &  R. H &  ID &  IDvsType  & Type &  R. W   \\
    \cmidrule{2-9}
    F   & 1.00$\pm 0.00$  & 0.95$\pm 0.07$  & 0.94$\pm 0.07$  & 0.99$\pm 0.02$  & 1.00$\pm 0.03$  & 1.00$\pm 0.00$  & 1.00$\pm 0.00$  & 0.99$\pm 0.02$ \\
 F+BD   & 1.00$\pm 0.00$  & 0.96$\pm 0.05$  & 0.95$\pm 0.06$  & 0.99$\pm 0.03$  & 1.00$\pm 0.00$  & 1.00$\pm 0.00$  & 1.00$\pm 0.00$  & 0.99$\pm 0.04$ \\
    M   & 1.00$\pm 0.00$  & 0.94$\pm 0.07$  & 0.93$\pm 0.07$  & 0.97$\pm 0.07$  & 1.00$\pm 0.00$  & 1.00$\pm 0.00$  & 1.00$\pm 0.00$  & 0.97$\pm 0.08$ \\
 M+BD   & 1.00$\pm 0.00$  & 0.95$\pm 0.07$  & 0.93$\pm 0.07$  & 0.97$\pm 0.07$  & 1.00$\pm 0.03$  & 1.00$\pm 0.00$  & 1.00$\pm 0.00$  & 0.97$\pm 0.08$ \\
   PM   & 1.00$\pm 0.00$  & 0.74$\pm 0.39$  & 0.92$\pm 0.06$  & 0.80$\pm 0.17$  & 1.00$\pm 0.00$  & 1.00$\pm 0.00$  & 1.00$\pm 0.00$  & 0.80$\pm 0.19$ \\
PM+BD   & 1.00$\pm 0.00$  & 0.86$\pm 0.25$  & 0.90$\pm 0.09$  & 0.83$\pm 0.19$  & 1.00$\pm 0.00$  & 1.00$\pm 0.00$  & 1.00$\pm 0.00$  & 0.81$\pm 0.19$ \\
    \midrule
     &\multicolumn{8}{c}{\makecell[tc]{Sampling Generation \\\bf Prompt Consistency Metrics}}\\
     {}   & Num. R. & Total Area &  R. Area &  R. H &  ID &  IDvsType  & Type &  R. W    \\
    \cmidrule{2-9}
 F   & 1.00$\pm 0.00$  & 0.95$\pm 0.06$  & 0.95$\pm 0.06$  & 0.99$\pm 0.03$  & 1.00$\pm 0.00$  & 1.00$\pm 0.00$  & 1.00$\pm 0.00$  & 0.99$\pm 0.02$ \\
 F+BD   & 1.00$\pm 0.00$  & 0.96$\pm 0.06$  & 0.95$\pm 0.06$  & 0.99$\pm 0.03$  & 1.00$\pm 0.00$  & 1.00$\pm 0.00$  & 1.00$\pm 0.00$  & 0.99$\pm 0.03$ \\
    M   & 1.00$\pm 0.00$  & 0.95$\pm 0.07$  & 0.93$\pm 0.08$  & 0.97$\pm 0.07$  & 1.00$\pm 0.00$  & 1.00$\pm 0.00$  & 1.00$\pm 0.00$  & 0.96$\pm 0.09$ \\
 M+BD   & 1.00$\pm 0.00$  & 0.95$\pm 0.06$  & 0.93$\pm 0.07$  & 0.96$\pm 0.07$  & 1.00$\pm 0.02$  & 1.00$\pm 0.00$  & 1.00$\pm 0.00$  & 0.97$\pm 0.08$ \\
   PM   & 1.00$\pm 0.00$  & 0.75$\pm 0.37$  & 0.91$\pm 0.07$  & 0.78$\pm 0.17$  & 1.00$\pm 0.01$  & 1.00$\pm 0.00$  & 1.00$\pm 0.00$  & 0.77$\pm 0.20$ \\
PM+BD   & 1.00$\pm 0.00$  & 0.86$\pm 0.26$  & 0.90$\pm 0.08$  & 0.79$\pm 0.19$  & 1.00$\pm 0.00$  & 1.00$\pm 0.00$  & 1.00$\pm 0.00$  & 0.79$\pm 0.19$ \\
    \bottomrule
\end{tabular}}
    \caption{Evaluation of \textbf{prompt consistency metrics} on generated results from \textbf{Specific Prompt}. Top half are results on single greedy generations from 1000 prompts from 1000 different test set floorplans; bottom half are results on sampled generations on 100 different prompt with 20 generations from each prompt.}
    \label{tab:detailed_eval}
\end{table}

\newpage
\subsection{Generation Prompt: Total Area}
\label{appendix:gen_procthor_total_area}

\begin{table}[h]
    \centering
    \small\begin{tabular}{lccccccc}
    \toprule
       &\multicolumn{7}{c}{\makecell[tc]{ Greedy Generation \\\bf Self Consistency Metrics}}\\
{}  &  P. Area  &  Overlap  &  R. Count  &  R. H  &  ID  &  R. W  &  Total Area  \\
    \cmidrule{2-8}
     F    & 0.79$\pm 0.13$  & 0.07$\pm 0.26$  & 0.96  & 1.00$\pm 0.01$  & 1.00  & 1.00$\pm 0.01$  & 0.97$\pm 0.07$ \\
  F+BD    & 0.83$\pm 0.09$  & 0.19$\pm 0.39$  & 0.97  & 1.00$\pm 0.02$  & 1.00  & 0.99$\pm 0.03$  & 0.97$\pm 0.09$ \\
     M    & 0.92$\pm 0.06$  & 0.09$\pm 0.28$  & 1.00  & 1.00$\pm 0.01$  & 1.00  & 1.00$\pm 0.01$  & 0.99$\pm 0.02$ \\
  M+BD    & 0.91$\pm 0.06$  & 0.14$\pm 0.35$  & 1.00  & 1.00$\pm 0.01$  & 1.00  & 1.00$\pm 0.01$  & 1.00$\pm 0.02$ \\
    PM    & 0.93$\pm 0.05$  & 0.14$\pm 0.35$  & 0.95  & 1.00$\pm 0.01$  & 1.00  & 1.00$\pm 0.01$  & 0.99$\pm 0.04$ \\
 PM+BD    & 0.90$\pm 0.07$  & 0.13$\pm 0.34$  & 0.98  & 1.00$\pm 0.01$  & 1.00  & 1.00$\pm 0.01$  & 0.98$\pm 0.12$ \\
    \midrule
     &\multicolumn{7}{c}{\makecell[tc]{Sampling Generation \\\bf Self Consistency Metrics}}\\
     {}  &  P. Area  &  Overlap  &  R. Count  &  R. H  &  ID  &  R. W  &  Total Area  \\
    \cmidrule{2-8}
              F   & 0.78$\pm 0.13$  & 0.14$\pm 0.34$  & 0.96  & 0.99$\pm 0.02$  & 1.00  & 0.99$\pm 0.02$  & 0.98$\pm 0.06$ \\
 F+BD   & 0.81$\pm 0.10$  & 0.26$\pm 0.44$  & 0.96  & 0.99$\pm 0.02$  & 1.00  & 0.99$\pm 0.02$  & 0.96$\pm 0.15$ \\
    M    & 0.90$\pm 0.06$  & 0.14$\pm 0.35$  & 1.00  & 1.00$\pm 0.01$  & 1.00  & 1.00$\pm 0.01$  & 1.00$\pm 0.03$ \\
 M+BD   & 0.90$\pm 0.06$  & 0.15$\pm 0.35$  & 1.00  & 1.00$\pm 0.01$  & 1.00  & 1.00$\pm 0.01$  & 0.99$\pm 0.03$ \\
   PM   & 0.92$\pm 0.06$  & 0.18$\pm 0.38$  & 0.97  & 1.00$\pm 0.01$  & 1.00  & 1.00$\pm 0.01$  & 0.99$\pm 0.03$ \\
PM+BD   & 0.88$\pm 0.07$  & 0.15$\pm 0.35$  & 0.96  & 1.00$\pm 0.01$  & 1.00  & 1.00$\pm 0.01$  & 0.96$\pm 0.18$ \\
    \bottomrule
\end{tabular}
    \caption{Evaluation of \textbf{self consistency metrics} on generated results from \textbf{Specific Prompt} (Section \ref{subsec:prompt_strat}). Top half are results on single greedy generations from 1000 prompts from 1000 different test set floorplans; bottom half are results on sampled generations on 100 different prompt with 20 generations from each prompt.}
    \label{tab:detailed_eval}
\end{table}

\begin{table}[h]
    \centering
    \small\begin{tabular}{lcccccccc}
    \toprule
       &\multicolumn{8}{c}{\makecell[tc]{ Greedy Generation \\\bf Prompt Consistency Metrics}}\\
{}   & Num. R. & Total Area &  R. Area &  R. H &  ID &  IDvsType  & Type &  R. W   \\
    \cmidrule{2-9}
       F  & 1.00$\pm 0.00$  & 0.83$\pm 0.14$  & -  & -  & -  & -  & -  & - \\
 F+BD  & 1.00$\pm 0.00$  & 0.87$\pm 0.12$  & -  & -  & -  & -  & -  & - \\
    M  & 1.00$\pm 0.00$  & 0.94$\pm 0.06$  & -  & -  & -  & -  & -  & - \\
 M+BD  & 1.00$\pm 0.00$  & 0.94$\pm 0.05$  & -  & -  & -  & -  & -  & - \\
   PM  & 1.00$\pm 0.00$  & 0.95$\pm 0.06$  & -  & -  & -  & -  & -  & - \\
PM+BD  & 1.00$\pm 0.00$  & 0.91$\pm 0.13$  & -  & -  & -  & -  & -  & - \\
    \midrule
     &\multicolumn{8}{c}{\makecell[tc]{Sampling Generation \\\bf Prompt Consistency Metrics}}\\
     {}   & Num. R. & Total Area &  R. Area &  R. H &  ID &  IDvsType  & Type &  R. W    \\
    \cmidrule{2-9}
     F  & 1.00$\pm 0.00$  & 0.82$\pm 0.15$  & -  & -  & -  & -  & -  & - \\
  F+BD  & 1.00$\pm 0.00$  & 0.84$\pm 0.16$  & -  & -  & -  & -  & -  & - \\
     M  & 1.00$\pm 0.00$  & 0.93$\pm 0.06$  & -  & -  & -  & -  & -  & - \\
  M+BD  & 1.00$\pm 0.00$  & 0.93$\pm 0.06$  & -  & -  & -  & -  & -  & - \\
    PM  & 1.00$\pm 0.00$  & 0.94$\pm 0.05$  & -  & -  & -  & -  & -  & - \\
 PM+BD  & 1.00$\pm 0.00$  & 0.88$\pm 0.18$  & -  & -  & -  & -  & -  & - \\
    \bottomrule
\end{tabular}
    \caption{Evaluation of \textbf{prompt consistency metrics} on generated results from \textbf{Total Area Prompt}. Top half are results on single greedy generations from 1000 prompts from 1000 different test set floorplans; bottom half are results on sampled generations on 100 different prompt with 20 generations from each prompt.}
    \label{tab:detailed_eval}
\end{table}

\newpage
\subsection{Generation Prompt: Partial Room Area}
\label{appendix:gen_procthor_partial_room_area}

\begin{table}[h]
    \centering
    \small\begin{tabular}{lccccccc}
    \toprule
       &\multicolumn{7}{c}{\makecell[tc]{ Greedy Generation \\\bf Self Consistency Metrics}}\\
{}  &  P. Area  &  Overlap  &  R. Count  &  R. H  &  ID  &  R. W  &  Total Area  \\
    \cmidrule{2-8}
         F  & 0.81$\pm 0.11$  & 0.03$\pm 0.17$  & 0.69  & 1.00$\pm 0.01$  & 1.00  & 1.00$\pm 0.02$  & 0.86$\pm 0.21$ \\
  F+BD  & 0.79$\pm 0.10$  & 0.17$\pm 0.37$  & 0.76  & 1.00$\pm 0.02$  & 1.00  & 0.99$\pm 0.03$  & 0.92$\pm 0.15$ \\
     M  & 0.90$\pm 0.07$  & 0.11$\pm 0.31$  & 0.94  & 0.99$\pm 0.02$  & 1.00  & 1.00$\pm 0.01$  & 0.97$\pm 0.10$ \\
  M+BD  & 0.90$\pm 0.06$  & 0.13$\pm 0.34$  & 0.99  & 1.00$\pm 0.01$  & 1.00  & 1.00$\pm 0.01$  & 0.99$\pm 0.04$ \\
    PM  & 0.92$\pm 0.06$  & 0.15$\pm 0.35$  & 0.96  & 1.00$\pm 0.01$  & 1.00  & 1.00$\pm 0.01$  & 0.97$\pm 0.07$ \\
 PM+BD  & 0.90$\pm 0.07$  & 0.14$\pm 0.34$  & 1.00  & 1.00$\pm 0.01$  & 1.00  & 1.00$\pm 0.01$  & 0.99$\pm 0.03$ \\
    \midrule
     &\multicolumn{7}{c}{\makecell[tc]{Sampling Generation \\\bf Self Consistency Metrics}}\\
     {}  &  P. Area  &  Overlap  &  R. Count  &  R. H  &  ID  &  R. W  &  Total Area  \\  
    \cmidrule{2-8}     
    F  & 0.79$\pm 0.13$  & 0.06$\pm 0.25$  & 0.75  & 0.99$\pm 0.02$  & 1.00  & 0.99$\pm 0.03$  & 0.87$\pm 0.22$ \\
  F+BD  & 0.78$\pm 0.11$  & 0.23$\pm 0.42$  & 0.81  & 0.99$\pm 0.02$  & 1.00  & 0.99$\pm 0.02$  & 0.94$\pm 0.13$ \\
     M  & 0.89$\pm 0.07$  & 0.14$\pm 0.34$  & 0.94  & 1.00$\pm 0.01$  & 1.00  & 1.00$\pm 0.02$  & 0.97$\pm 0.11$ \\
  M+BD  & 0.89$\pm 0.07$  & 0.14$\pm 0.35$  & 0.99  & 1.00$\pm 0.01$  & 1.00  & 1.00$\pm 0.01$  & 0.99$\pm 0.03$ \\
    PM  & 0.91$\pm 0.07$  & 0.18$\pm 0.39$  & 0.98  & 1.00$\pm 0.01$  & 1.00  & 1.00$\pm 0.01$  & 0.98$\pm 0.04$ \\
 PM+BD  & 0.88$\pm 0.07$  & 0.17$\pm 0.38$  & 1.00  & 1.00$\pm 0.01$  & 1.00  & 1.00$\pm 0.01$  & 0.99$\pm 0.04$ \\
    \bottomrule
\end{tabular}
    \caption{Evaluation of \textbf{self consistency metrics} on generated results from \textbf{Specific Prompt} (Section \ref{subsec:prompt_strat}). Top half are results on single greedy generations from 1000 prompts from 1000 different test set floorplans; bottom half are results on sampled generations on 100 different prompt with 20 generations from each prompt.}
    \label{tab:detailed_eval}
\end{table}

\begin{table}[h]
    \centering
    \resizebox{\textwidth}{!}{\small\begin{tabular}{lcccccccc}
    \toprule
       &\multicolumn{8}{c}{\makecell[tc]{ Greedy Generation \\\bf Prompt Consistency Metrics}}\\
{}   & Num. R. & Total Area &  R. Area &  R. H &  ID &  IDvsType  & Type &  R. W   \\
    \cmidrule{2-9}
          F  & 1.00$\pm 0.00$  & 0.48$\pm 0.40$  & 0.82$\pm 0.14$  & -  & 0.99$\pm 0.09$  & 1.00$\pm 0.02$  & 0.99$\pm 0.07$  & - \\
   F+BD  & 1.00$\pm 0.00$  & 0.74$\pm 0.21$  & 0.79$\pm 0.14$  & -  & 0.97$\pm 0.15$  & 1.00$\pm 0.00$  & 0.99$\pm 0.06$  & - \\
      M  & 1.00$\pm 0.00$  & 0.91$\pm 0.10$  & 0.80$\pm 0.35$  & -  & 0.99$\pm 0.07$  & 0.89$\pm 0.27$  & 0.99$\pm 0.09$  & - \\
   M+BD  & 1.00$\pm 0.00$  & 0.93$\pm 0.06$  & 0.87$\pm 0.13$  & -  & 1.00$\pm 0.05$  & 0.99$\pm 0.11$  & 1.00$\pm 0.00$  & - \\
     PM  & 1.00$\pm 0.00$  & 0.93$\pm 0.07$  & 0.91$\pm 0.11$  & -  & 0.99$\pm 0.05$  & 1.00$\pm 0.00$  & 1.00$\pm 0.04$  & - \\
  PM+BD  & 1.00$\pm 0.00$  & 0.92$\pm 0.07$  & 0.89$\pm 0.10$  & -  & 1.00$\pm 0.02$  & 1.00$\pm 0.00$  & 1.00$\pm 0.01$  & - \\
    \midrule
     &\multicolumn{8}{c}{\makecell[tc]{Sampling Generation \\\bf Prompt Consistency Metrics}}\\
     {}   & Num. R. & Total Area &  R. Area &  R. H &  ID &  IDvsType  & Type &  R. W    \\
    \cmidrule{2-9}
        F  & 1.00$\pm 0.02$  & 0.50$\pm 0.39$  & 0.80$\pm 0.15$  & -  & 0.99$\pm 0.07$  & 1.00$\pm 0.00$  & 1.00$\pm 0.06$  & - \\
 F+BD  & 1.00$\pm 0.00$  & 0.73$\pm 0.25$  & 0.75$\pm 0.23$  & -  & 0.96$\pm 0.16$  & 1.00$\pm 0.02$  & 0.99$\pm 0.07$  & - \\
    M  & 1.00$\pm 0.00$  & 0.90$\pm 0.12$  & 0.80$\pm 0.26$  & -  & 0.99$\pm 0.07$  & 0.88$\pm 0.29$  & 0.99$\pm 0.07$  & - \\
 M+BD  & 1.00$\pm 0.00$  & 0.93$\pm 0.07$  & 0.87$\pm 0.14$  & -  & 0.99$\pm 0.06$  & 0.98$\pm 0.15$  & 1.00$\pm 0.00$  & - \\
   PM  & 1.00$\pm 0.00$  & 0.94$\pm 0.06$  & 0.90$\pm 0.11$  & -  & 1.00$\pm 0.03$  & 1.00$\pm 0.00$  & 1.00$\pm 0.01$  & - \\
PM+BD  & 1.00$\pm 0.00$  & 0.90$\pm 0.08$  & 0.87$\pm 0.13$  & -  & 1.00$\pm 0.02$  & 1.00$\pm 0.00$  & 1.00$\pm 0.01$  & - \\
    \bottomrule
\end{tabular}}
    \caption{Evaluation of \textbf{prompt consistency metrics} on generated results from \textbf{Partial Room Area Prompt}. Top half are results on single greedy generations from 1000 prompts from 1000 different test set floorplans; bottom half are results on sampled generations on 100 different prompt with 20 generations from each prompt.}
    \label{tab:detailed_eval}
\end{table}

\newpage
\subsection{Generation Prompt: All Room Area}
\label{appendix:gen_procthor_all_room_area}

\begin{table}[h]
    \centering
    \small\begin{tabular}{lccccccc}
    \toprule
       &\multicolumn{7}{c}{\makecell[tc]{ Greedy Generation \\\bf Self Consistency Metrics}}\\
{}  &  P. Area  &  Overlap  &  R. Count  &  R. H  &  ID  &  R. W  &  Total Area  \\
    \cmidrule{2-8}
     F  & 0.82$\pm 0.10$  & 0.16$\pm 0.37$  & 1.00  & 1.00$\pm 0.02$  & 1.00  & 1.00$\pm 0.01$  & 0.93$\pm 0.17$ \\
  F+BD  & 0.76$\pm 0.11$  & 0.25$\pm 0.43$  & 1.00  & 0.99$\pm 0.02$  & 1.00  & 0.99$\pm 0.02$  & 0.90$\pm 0.20$ \\
     M  & 0.90$\pm 0.07$  & 0.12$\pm 0.33$  & 1.00  & 0.99$\pm 0.02$  & 1.00  & 0.99$\pm 0.01$  & 0.99$\pm 0.03$ \\
  M+BD  & 0.91$\pm 0.06$  & 0.14$\pm 0.35$  & 1.00  & 1.00$\pm 0.01$  & 1.00  & 1.00$\pm 0.01$  & 0.99$\pm 0.02$ \\
    PM  & 0.91$\pm 0.06$  & 0.14$\pm 0.35$  & 1.00  & 1.00$\pm 0.01$  & 1.00  & 1.00$\pm 0.01$  & 0.97$\pm 0.09$ \\
 PM+BD  & 0.90$\pm 0.07$  & 0.19$\pm 0.39$  & 1.00  & 1.00$\pm 0.01$  & 1.00  & 1.00$\pm 0.01$  & 0.99$\pm 0.05$ \\
    \midrule
     &\multicolumn{7}{c}{\makecell[tc]{Sampling Generation \\\bf Self Consistency Metrics}}\\
     {}  &  P. Area  &  Overlap  &  R. Count  &  R. H  &  ID  &  R. W  &  Total Area  \\  
    \cmidrule{2-8}
     F  & 0.80$\pm 0.10$  & 0.18$\pm 0.39$  & 0.99  & 0.99$\pm 0.02$  & 1.00  & 0.99$\pm 0.04$  & 0.95$\pm 0.14$ \\
  F+BD  & 0.76$\pm 0.11$  & 0.31$\pm 0.46$  & 1.00  & 0.99$\pm 0.02$  & 1.00  & 0.99$\pm 0.03$  & 0.93$\pm 0.15$ \\
     M  & 0.89$\pm 0.07$  & 0.17$\pm 0.37$  & 1.00  & 1.00$\pm 0.01$  & 1.00  & 0.99$\pm 0.02$  & 0.99$\pm 0.03$ \\
  M+BD  & 0.90$\pm 0.07$  & 0.15$\pm 0.36$  & 1.00  & 1.00$\pm 0.01$  & 1.00  & 1.00$\pm 0.01$  & 0.99$\pm 0.04$ \\
    PM  & 0.91$\pm 0.07$  & 0.18$\pm 0.38$  & 1.00  & 1.00$\pm 0.01$  & 1.00  & 1.00$\pm 0.01$  & 0.96$\pm 0.10$ \\
 PM+BD  & 0.88$\pm 0.07$  & 0.19$\pm 0.40$  & 1.00  & 1.00$\pm 0.01$  & 1.00  & 1.00$\pm 0.01$  & 1.00$\pm 0.00$ \\
    \bottomrule
\end{tabular}
    \caption{Evaluation of \textbf{self consistency metrics} on generated results from \textbf{Specific Prompt} (Section \ref{subsec:prompt_strat}). Top half are results on single greedy generations from 1000 prompts from 1000 different test set floorplans; bottom half are results on sampled generations on 100 different prompt with 20 generations from each prompt.}
    \label{tab:detailed_eval}
\end{table}

\begin{table}[h]
    \centering
    \resizebox{\textwidth}{!}{\small\begin{tabular}{lcccccccc}
    \toprule
       &\multicolumn{8}{c}{\makecell[tc]{ Greedy Generation \\\bf Prompt Consistency Metrics}}\\
{}   & Num. R. & Total Area &  R. Area &  R. H &  ID &  IDvsType  & Type &  R. W   \\
    \cmidrule{2-9}
        F  & 1.00$\pm 0.00$  & -  & 0.81$\pm 0.10$  & -  & 1.00$\pm 0.06$  & 1.00$\pm 0.00$  & 1.00$\pm 0.06$  & - \\
 F+BD  & 1.00$\pm 0.00$  & -  & 0.76$\pm 0.11$  & -  & 1.00$\pm 0.00$  & 1.00$\pm 0.00$  & 1.00$\pm 0.00$  & - \\
    M  & 1.00$\pm 0.00$  & -  & 0.90$\pm 0.07$  & -  & 1.00$\pm 0.00$  & 1.00$\pm 0.00$  & 1.00$\pm 0.00$  & - \\
 M+BD  & 1.00$\pm 0.00$  & -  & 0.91$\pm 0.06$  & -  & 1.00$\pm 0.01$  & 1.00$\pm 0.00$  & 1.00$\pm 0.00$  & - \\
   PM  & 1.00$\pm 0.00$  & -  & 0.91$\pm 0.06$  & -  & 1.00$\pm 0.00$  & 1.00$\pm 0.00$  & 1.00$\pm 0.00$  & - \\
PM+BD  & 1.00$\pm 0.00$  & -  & 0.90$\pm 0.07$  & -  & 1.00$\pm 0.00$  & 1.00$\pm 0.00$  & 1.00$\pm 0.00$  & - \\
    \midrule
     &\multicolumn{8}{c}{\makecell[tc]{Sampling Generation \\\bf Prompt Consistency Metrics}}\\
     {}   & Num. R. & Total Area &  R. Area &  R. H &  ID &  IDvsType  & Type &  R. W    \\
    \cmidrule{2-9}
        F  & 1.00$\pm 0.00$  & -  & 0.79$\pm 0.10$  & -  & 0.99$\pm 0.09$  & 1.00$\pm 0.00$  & 0.99$\pm 0.09$  & - \\
 F+BD  & 1.00$\pm 0.00$  & -  & 0.76$\pm 0.11$  & -  & 1.00$\pm 0.00$  & 1.00$\pm 0.00$  & 1.00$\pm 0.00$  & - \\
    M  & 1.00$\pm 0.00$  & -  & 0.89$\pm 0.07$  & -  & 1.00$\pm 0.00$  & 1.00$\pm 0.00$  & 1.00$\pm 0.00$  & - \\
 M+BD  & 1.00$\pm 0.00$  & -  & 0.90$\pm 0.07$  & -  & 1.00$\pm 0.04$  & 1.00$\pm 0.00$  & 1.00$\pm 0.00$  & - \\
   PM  & 1.00$\pm 0.00$  & -  & 0.91$\pm 0.07$  & -  & 1.00$\pm 0.01$  & 1.00$\pm 0.00$  & 1.00$\pm 0.01$  & - \\
PM+BD  & 1.00$\pm 0.00$  & -  & 0.88$\pm 0.07$  & -  & 1.00$\pm 0.02$  & 1.00$\pm 0.00$  & 1.00$\pm 0.00$  & - \\
    \bottomrule
\end{tabular}}
    \caption{Evaluation of \textbf{prompt consistency metrics} on generated results from \textbf{All Room Area Prompt}. Top half are results on single greedy generations from 1000 prompts from 1000 different test set floorplans; bottom half are results on sampled generations on 100 different prompt with 20 generations from each prompt.}
    \label{tab:detailed_eval}
\end{table}

\newpage
\section{Compatibility Metrics on ProcTHOR}
\label{appendix:procthor_compatibility}
We further run compatibility evaluations on all of the Bubble Diagram model variants to test 
how well bubble diagrams are followed. The results are presented here:

\begin{table}[h]
\centering
\small\begin{tabular}{ccccccc}
\toprule
Model & Generation & \multicolumn{5}{c}{Number of Rooms}\\
\cmidrule{3-7}
Variant &  Prompt   & 3   & 4  & 5 & 6 & 7  \\
\midrule
F+BD        & Specific   & 0.15 $\pm$0.41 & 0.33 $\pm$0.58 & 1.35 $\pm$1.34 & 2.24 $\pm$1.76 & 4.23 $\pm$1.89 \\
            & Total Area    & 0.19 $\pm$0.40 & 0.62 $\pm$0.78 & 3.48 $\pm$1.83 & 3.41 $\pm$1.50 & 6.23 $\pm$2.20   \\
            & Partial Room Area &  3.80 $\pm$1.20  & 5.73 $\pm$2.03  & 8.19 $\pm$2.72 & 10.15 $\pm$2.88  & 11.42 $\pm$4.52 \\
            & All Room Area   & 5.91 $\pm$0.29 & 9.75 $\pm$0.54 & 13.79 $\pm$1.15 & 18.09 $\pm$1.71 & 22.35 $\pm$2.24  \\
\midrule
M+BD &Specific   &  0.16 $\pm$0.45 &  0.47 $\pm$0.67 &  2.06 $\pm$1.33 &  2.85 $\pm$1.79 &  5.23 $\pm$2.14 \\
        &Total Area   & 0.20 $\pm$0.40 & 0.54 $\pm$0.73 & 2.40 $\pm$1.03 & 3.09 $\pm$2.07 & 6.26 $\pm$2.03  \\
        &Partial Room Area    & 3.85 $\pm$1.11 & 6.35 $\pm$1.95 & 8.44 $\pm$2.48 & 11.56 $\pm$3.42 & 14.13 $\pm$3.90   \\
        &All Room Area   &5.91 $\pm$0.29 & 9.62 $\pm$0.56 & 12.90 $\pm$1.27 & 17.48 $\pm$1.82 & 22.13 $\pm$1.94  \\
\midrule
PM+BD &Specific  & 0.15 $\pm$0.39 & 0.58 $\pm$0.96 & 2.42 $\pm$1.33 & 2.97 $\pm$1.62 & 5.97 $\pm$1.25 \\
        &Total Area   &  0.34 $\pm$0.93 & 0.46 $\pm$0.84 & 2.08 $\pm$1.29 & 2.65 $\pm$1.87 & 5.84 $\pm$1.61   \\
        &Partial Room Area   & 3.87 $\pm$1.13 & 6.42 $\pm$1.97 & 8.99 $\pm$2.67 & 12.06 $\pm$3.05 & 13.68 $\pm$3.71    \\
        &All Room Area   & 5.88 $\pm$0.33 & 9.77 $\pm$0.49 & 13.17 $\pm$1.23 & 18.56 $\pm$1.58 & 21.97 $\pm$2.36   \\
\bottomrule
\end{tabular}
    \caption{Compatibility ($\downarrow$) score on ProcTHOR experiments. The best score in each category (by number of rooms in a floor plan) is bolded, and the 2nd best score has an asterisk.}
    \label{tab:procthor_ged_all}
\end{table}

\section{Self and Prompt Consistency Metrics on RPLAN}
\label{appendix:rplan_consistency_metrics}
This section contains the full self- and prompt-consistency metrics evaluated on the 4 models (Section \ref{subsec:model_variants}) trained on the converted RPLAN dataset.

Similiar to the ProcTHOR-trained model experiments, we run the 4 different generation prompts on each of the RPLAN-trained model variants, resulting in 16 different experiments. They are organized by generation prompt and shown here.

\subsection{Generation Prompt: Specific}
\label{appendix:gen_rplan_specific}
\begin{table}[h]
    \centering
    \small\begin{tabular}{lccccccc}
    \toprule
       &\multicolumn{7}{c}{\bf Self Consistency Metrics}\\
{}  &  P. Area  &  Overlap  &  R. Count  &  R. H  &  ID  &  R. W  &  Total Area  \\
    \cmidrule{2-8}
    5-Room    & 0.89$\pm 0.31$  & 0.34$\pm 0.47$  & 0.40  & 1.00$\pm 0.01$  & 1.00  & 0.98$\pm 0.03$  & 0.92$\pm 0.08$ \\
    6-Room    & 0.93$\pm 0.05$  & 0.36$\pm 0.48$  & 0.99  & 0.99$\pm 0.01$  & 1.00  & 0.97$\pm 0.02$  & 0.98$\pm 0.06$ \\
    7-Room    & 0.93$\pm 0.04$  & 0.42$\pm 0.49$  & 0.97  & 0.99$\pm 0.01$  & 1.00  & 0.98$\pm 0.02$  & 0.98$\pm 0.05$ \\
    8-Room    & 0.93$\pm 0.07$  & 0.62$\pm 0.49$  & 0.89  & 0.99$\pm 0.01$  & 1.00  & 0.98$\pm 0.02$  & 0.95$\pm 0.08$ \\
    \bottomrule
\end{tabular}
    \caption{\textbf{Self consistency} metrics on RPLAN generations using \textbf{Specific} generation prompts.}
    \label{tab:rplan_metrics_fullprompt_sc}
\end{table}

\begin{table}[h]
    \centering
    \resizebox{\textwidth}{!}{\small\begin{tabular}{lcccccccc}
    \toprule
       &\multicolumn{8}{c}{\bf Prompt Consistency Metrics}\\
{}   & Num. R. & Total Area &  R. Area &  R. H &  ID &  IDvsType  & Type &  R. W   \\
    \cmidrule{2-9}
     5-Room   & 1.00$\pm 0.00$  & 0.92$\pm 0.06$  & 0.93$\pm 0.09$  & 0.94$\pm 0.11$  & 1.00$\pm 0.00$  & 1.00$\pm 0.00$  & 1.00$\pm 0.00$  & 0.95$\pm 0.09$ \\
 6-Room   & 1.00$\pm 0.00$  & 0.94$\pm 0.07$  & 0.94$\pm 0.06$  & 0.92$\pm 0.23$  & 1.00$\pm 0.05$  & 1.00$\pm 0.00$  & 1.00$\pm 0.00$  & 0.94$\pm 0.14$ \\
 7-Room   & 1.00$\pm 0.00$  & 0.94$\pm 0.06$  & 0.94$\pm 0.06$  & 0.93$\pm 0.17$  & 1.00$\pm 0.02$  & 1.00$\pm 0.00$  & 1.00$\pm 0.00$  & 0.94$\pm 0.29$ \\
 8-Room   & 1.00$\pm 0.00$  & 0.92$\pm 0.08$  & 0.94$\pm 0.07$  & 0.95$\pm 0.10$  & 0.99$\pm 0.06$  & 1.00$\pm 0.00$  & 1.00$\pm 0.00$  & 0.96$\pm 0.09$ \\
    \bottomrule
\end{tabular}}
    \caption{\textbf{Prompt consistency} metrics on RPLAN generations using \textbf{Specific} generation prompts.}
    \label{tab:rplan_metrics_fullprompt_pc}
\end{table}

\newpage
\subsection{Generation Prompt: Total Area}
\label{appendix:gen_rplan_total_area}
\begin{table}[h]
    \centering
    \small\begin{tabular}{lccccccc}
    \toprule
       &\multicolumn{7}{c}{\bf Self Consistency Metrics}\\
{}  &  P. Area  &  Overlap  &  R. Count  &  R. H  &  ID  &  R. W  &  Total Area  \\
    \cmidrule{2-8}
    5-Room   & 0.89$\pm 0.41$  & 0.31$\pm 0.46$  & 0.19  & 1.00$\pm 0.01$  & 1.00  & 0.98$\pm 0.03$  & 0.95$\pm 0.06$ \\
 6-Room    & 0.93$\pm 0.05$  & 0.32$\pm 0.47$  & 1.00  & 0.99$\pm 0.01$  & 1.00  & 0.98$\pm 0.02$  & 0.99$\pm 0.03$ \\
 7-Room  & 0.94$\pm 0.03$  & 0.38$\pm 0.48$  & 1.00  & 0.99$\pm 0.01$  & 1.00  & 0.98$\pm 0.02$  & 0.99$\pm 0.02$ \\
 8-Room   & 0.82$\pm 3.50$  & 0.53$\pm 0.50$  & 0.89  & 0.99$\pm 0.01$  & 1.00  & 0.98$\pm 0.02$  & 0.97$\pm 0.07$ \\
    \bottomrule
\end{tabular}
    \caption{\textbf{Self consistency} metrics on RPLAN generations using \textbf{Total Area} generation prompts.}
    \label{tab:rplan_metrics_totalarea_sc}
\end{table}

\begin{table}[h]
    \centering
    \small\begin{tabular}{lcccccccc}
    \toprule
       &\multicolumn{8}{c}{\bf Prompt Consistency Metrics}\\
{}   & Num. R. & Total Area &  R. Area &  R. H &  ID &  IDvsType  & Type &  R. W   \\
    \cmidrule{2-9}
     5-Room   & -  & 0.92$\pm 0.07$  & -  & -  & -  & -  & -  & - \\
    6-Room & -  & 0.94$\pm 0.06$  & -  & -  & -  & -  & -  & - \\
    7-Room & -  & 0.94$\pm 0.06$  & -  & -  & -  & -  & -  & - \\
    8-Room  & -  & 0.93$\pm 0.07$  & -  & -  & -  & -  & -  & - \\
    \bottomrule
\end{tabular}
    \caption{\textbf{Prompt consistency} metrics on RPLAN generations using \textbf{Total Area} generation prompts.}
    \label{tab:rplan_metrics_totalarea_pc}
\end{table}

\subsection{Generation Prompt: Partial Room Area}
\label{appendix:gen_rplan_partial_room_area}
\begin{table}[h]
    \centering
    \small\begin{tabular}{lccccccc}
    \toprule
       &\multicolumn{7}{c}{\bf Self Consistency Metrics}\\
{}  &  P. Area  &  Overlap  &  R. Count  &  R. H  &  ID  &  R. W  &  Total Area  \\
    \cmidrule{2-8}
    5-Room   & 0.77$\pm 1.93$  & 0.25$\pm 0.44$  & 0.49  & 1.00$\pm 0.01$  & 1.00  & 0.97$\pm 0.10$  & 0.95$\pm 0.08$ \\
6-Room   & 0.93$\pm 0.07$  & 0.34$\pm 0.47$  & 0.98  & 0.99$\pm 0.04$  & 1.00  & 0.97$\pm 0.02$  & 0.97$\pm 0.08$ \\
7-Room  & 0.93$\pm 0.04$  & 0.40$\pm 0.49$  & 0.96  & 0.99$\pm 0.02$  & 1.00  & 0.98$\pm 0.02$  & 0.96$\pm 0.07$ \\
8-Room   & 0.91$\pm 0.34$  & 0.54$\pm 0.50$  & 0.73  & 0.99$\pm 0.02$  & 1.00  & 0.97$\pm 0.02$  & 0.93$\pm 0.11$ \\
    \bottomrule
\end{tabular}
    \caption{\textbf{Self consistency} metrics on RPLAN generations using \textbf{Partial Room Area} generation prompts.}
    \label{tab:rplan_metrics_partialroomarea_sc}
\end{table}

\begin{table}[hb!]
    \centering
    \resizebox{\textwidth}{!}{\small\begin{tabular}{lcccccccc}
    \toprule
       &\multicolumn{8}{c}{\bf Prompt Consistency Metrics}\\
 & Num. R. & Total Area &  R. Area &  R. H &  ID &  IDvsType  & Type &  R. W   \\
    \cmidrule{2-9}
     5-Room   & -  & 0.92$\pm 0.08$  & 0.91$\pm 0.14$  & -  & 1.00$\pm 0.05$  & 1.00$\pm 0.04$  & 1.00$\pm 0.00$  & - \\
6-Room   & -  & 0.93$\pm 0.08$  & 0.86$\pm 0.72$  & -  & 0.99$\pm 0.06$  & 0.99$\pm 0.07$  & 1.00$\pm 0.02$  & - \\
7-Room   & -  & 0.91$\pm 0.09$  & 0.88$\pm 0.19$  & -  & 1.00$\pm 0.03$  & 0.99$\pm 0.06$  & 1.00$\pm 0.01$  & - \\
8-Room   & -  & 0.89$\pm 0.11$  & 0.73$\pm 0.48$  & -  & 0.98$\pm 0.09$  & 0.94$\pm 0.13$  & 0.99$\pm 0.05$  & - \\
    \bottomrule
\end{tabular}}
    \caption{\textbf{Prompt consistency} metrics on RPLAN generations using \textbf{Partial Room Area} generation prompts.}
    \label{tab:rplan_metrics_partialroomarea_pc}
\end{table}

\newpage
\subsection{Generation Prompt: All Room Area}
\label{appendix:gen_rplan_all_room_area}
\begin{table}[h]
    \centering
    \small\begin{tabular}{lccccccc}
    \toprule
       &\multicolumn{7}{c}{\bf Self Consistency Metrics}\\
{}  &  P. Area  &  Overlap  &  R. Count  &  R. H  &  ID  &  R. W  &  Total Area  \\
    \cmidrule{2-8}
    5-Room  & 0.90$\pm 0.16$  & 0.37$\pm 0.48$  & 0.97  & 1.00$\pm 0.01$  & 1.00  & 0.98$\pm 0.02$  & 0.98$\pm 0.05$ \\
6-Room  & 0.93$\pm 0.06$  & 0.31$\pm 0.46$  & 1.00  & 0.99$\pm 0.01$  & 1.00  & 0.97$\pm 0.03$  & 0.95$\pm 0.04$ \\
7-Room  & 0.92$\pm 0.15$  & 0.38$\pm 0.49$  & 0.93  & 0.99$\pm 0.02$  & 1.00  & 0.98$\pm 0.02$  & 0.98$\pm 0.03$ \\
8-Room   & 0.93$\pm 0.10$  & 0.61$\pm 0.49$  & 0.96  & 0.99$\pm 0.02$  & 1.00  & 0.98$\pm 0.02$  & 0.91$\pm 0.07$ \\
    \bottomrule
\end{tabular}
    \caption{\textbf{Self consistency} metrics on RPLAN generations using \textbf{All Room Area} generation prompts.}
    \label{tab:rplan_metrics_allroomarea_sc}
\end{table}

\begin{table}[h]
    \centering
    \small\begin{tabular}{lcccccccc}
    \toprule
       &\multicolumn{8}{c}{\bf Prompt Consistency Metrics}\\
{}   & Num. R. & Total Area &  R. Area &  R. H &  ID &  IDvsType  & Type &  R. W   \\
    \cmidrule{2-9}
     5-Room   & -  & -  & 0.92$\pm 0.04$  & -  & 1.00$\pm 0.01$  & 1.00$\pm 0.00$  & 1.00$\pm 0.00$  & - \\
 6-Room   & -  & -  & 0.93$\pm 0.05$  & -  & 1.00$\pm 0.02$  & 1.00$\pm 0.00$  & 1.00$\pm 0.00$  & - \\
 7-Room   & -  & -  & 0.93$\pm 0.07$  & -  & 1.00$\pm 0.02$  & 1.00$\pm 0.00$  & 1.00$\pm 0.00$  & - \\
 8-Room   & -  & -  & 0.93$\pm 0.05$  & -  & 0.99$\pm 0.04$  & 1.00$\pm 0.00$  & 1.00$\pm 0.00$  & - \\
    \bottomrule
\end{tabular}
    \caption{\textbf{Prompt consistency} metrics on RPLAN generations using \textbf{All Room Area} generation prompts.}
    \label{tab:rplan_metrics_allroomarea_pc}
\end{table}

\newpage
\section{Generation Example}
\label{appendix:gen_example}
\begin{figure}[h]
    \centering
    \includegraphics[width=0.5\linewidth]{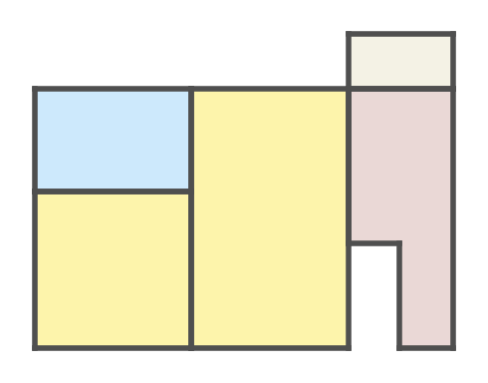}
    \caption{This Floor plan is generated by the string output below}
    \label{fig:enter-label}
\end{figure}

\begin{verbatim}
{
  "room_count": 5,
  "total_area": 183.6,
  "room_types": [
    "Bedroom",
    "Bathroom",
    "Bedroom",
    "Kitchen",
    "LivingRoom"
  ],
  "rooms": [
    {
      "area": 41.3,
      "floor_polygon": [
        {
          "x": 0,
          "z": 0
        },
        {
          "x": 0,
          "z": 6.4
        },
        {
          "x": 6.4,
          "z": 6.4
        },
        {
          "x": 6.4,
          "z": 0
        }
      ],
      "height": 6.4,
      "id": "room|4",
      "is_regular": 1,
      "room_type": "Bedroom",
      "width": 6.4
    },
    {
      "area": 27.5,
      "floor_polygon": [
        {
          "x": 0,
          "z": 6.4
        },
        {
          "x": 0,
          "z": 10.6
        },
        {
          "x": 6.4,
          "z": 10.6
        },
        {
          "x": 6.4,
          "z": 6.4
        }
      ],
      "height": 4.3,
      "id": "room|5",
      "is_regular": 1,
      "room_type": "Bathroom",
      "width": 6.4
    },
    {
      "area": 59.7,
      "floor_polygon": [
        {
          "x": 6.4,
          "z": 0
        },
        {
          "x": 6.4,
          "z": 10.6
        },
        {
          "x": 12.8,
          "z": 10.6
        },
        {
          "x": 12.8,
          "z": 0
        }
      ],
      "height": 10.6,
      "id": "room|6",
      "is_regular": 0,
      "room_type": "Bedroom",
      "width": 6.4
    },
    {
      "area": 36.7,
      "floor_polygon": [
        {
          "x": 12.8,
          "z": 4.3
        },
        {
          "x": 12.8,
          "z": 10.6
        },
        {
          "x": 17.1,
          "z": 10.6
        },
        {
          "x": 17.1,
          "z": 0
        },
        {
          "x": 14.9,
          "z": 0
        },
        {
          "x": 14.9,
          "z": 4.3
        }
      ],
      "height": 10.6,
      "id": "room|8",
      "is_regular": 0,
      "room_type": "Kitchen",
      "width": 4.3
    },
    {
      "area": 18.4,
      "floor_polygon": [
        {
          "x": 12.8,
          "z": 10.6
        },
        {
          "x": 12.8,
          "z": 12.8
        },
        {
          "x": 17.1,
          "z": 12.8
        },
        {
          "x": 17.1,
          "z": 10.6
        }
      ],
      "height": 2.1,
      "id": "room|9",
      "is_regular": 1,
      "room_type": "LivingRoom",
      "width": 4.3
    }
  ]
}
\end{verbatim}